\documentclass[letterpaper,10pt]{article}
\usepackage[preprint]{neurips_2026}

\usepackage{hyperref}
\usepackage{url}
\usepackage{preamble}
\usepackage{data_macros}

\definecolor{cherryred}{HTML}{BF0140}
\definecolor{refblue}{HTML}{002170}
\hypersetup{
  colorlinks=true,
  linkcolor=cherryred,
  citecolor=cherryred,
  urlcolor=cherryred,
  pdftitle={Architecture Determines Observability of Transformers},
  pdfauthor={Thomas Carmichael},
  pdfsubject={Observability and signal engineering in autoregressive transformers},
  pdfkeywords={transformers, observability, activation monitoring,
    signal engineering, monitorability, confidence control,
    output confidence, probing, representation geometry,
    training dynamics, calibration, interpretability, language models}
}

\title{Architecture Determines Observability of Transformers}

\author{%
  Thomas Carmichael \\
  Independent Researcher \\
  \href{mailto:github@tdosmail.com}{\textcolor{black}{\texttt{github@tdosmail.com}}}
}

\makeatletter
\renewcommand{\@notice}{}
\makeatother

\begin{document}

\maketitle

\begin{abstract}
\setlength{\parskip}{0.6em plus 0.1em minus 0.1em}

Autoregressive transformers make confident errors that
output-confidence monitoring cannot catch. Activation monitors catch them
only when training leaves a decision-quality signal beyond what the
output already exposes.
This signal is an architectural property of the trained model,
fixed upstream of any monitor. Controlling for
output confidence removes $\confabsorbmean$\% of the raw
activation-probe signal on average across \nabsorbmodels{} models.
Raw probe signal is mostly output confidence, and
output-side readouts cannot recover the residual. What remains depends on architecture and training.
In Pythia's controlled training, both matched-width configurations
form the signal early. One preserves it through convergence while
another erases it as perplexity continues to improve. Capability and observability are not inherently in tension. Across independently trained families
this pattern persists, even as the collapse point shifts. Where the signal
survives, monitoring catches what confidence cannot. On downstream QA, a WikiText-trained
probe with no task-specific tuning catches about one in eight
confident errors that output-confidence monitoring misses, at a
20\% flag rate. These results establish signal engineering as a training-time
design axis alongside loss and capability. Architecture sets the conditions for observability,
and training determines what remains readable.
\end{abstract}

\section{Introduction}
\label{sec:introduction}

Activation monitors for trained models operate on the premise that the
target signal survives training~\citep{kramar2026building,zou2023representation}.
Hidden states encode signals not directly exposed by outputs, including
latent knowledge, geometric structure of truthful representations, and
verification of generated
reasoning~\citep{burns2023discovering,marks2024geometry,zhang2025reasoning}.
This paper tests two output-side signals: max-softmax probability and
trained final-layer predictors. A linear probe on frozen
mid-layer activations~\citep{alain2017understanding,hewitt2019designing,belinkov2022probing}
can read a different signal: per-token decision quality beyond what
output confidence explains. On some architectures, this signal is
strong and stable across seeds. On others, no measured layer
preserves a healthy-range, linearly readable signal.

Monitorability is therefore an architectural property before it is a
monitor-design problem. We make this property measurable by defining
observability: the confidence-controlled linear readability of
per-token decision quality from frozen mid-layer activations
(\cref{sec:method}). For this decision-quality target, much of what
raw probes find in hidden states is output confidence in disguise.
On GPT-2~124M, raw Spearman correlation between a linear probe and
per-token loss is $\rawspearman$, but only $\stdcontrol$ survives
after controlling for max-softmax and activation norm. Across
\nabsorbmodels{} models in \nfamilies{} families, controlling for
confidence removes on average $\confabsorbmean$\% of the raw signal.
Raw probe scores therefore substantially overstate what activations
reveal beyond output confidence.

\begin{figure}[t]
\centering
\includegraphics[width=\textwidth,alt={Two side-by-side panels showing
  layer-wise partial correlation across within-family configurations.
  Left panel, Llama: 1B rises to 0.28 through forward pass; 3B and
  8B stay flat near 0.05 to 0.10. Right panel, Pythia: six sizes
  (70M, 160M, 1B, 2.8B, 6.9B, 12B) show mid-to-late peaks between
  0.20 and 0.38; 410M and 1.4B, both at 24 layers and 16 heads, stay
  flat near 0.10 across all depths.}]{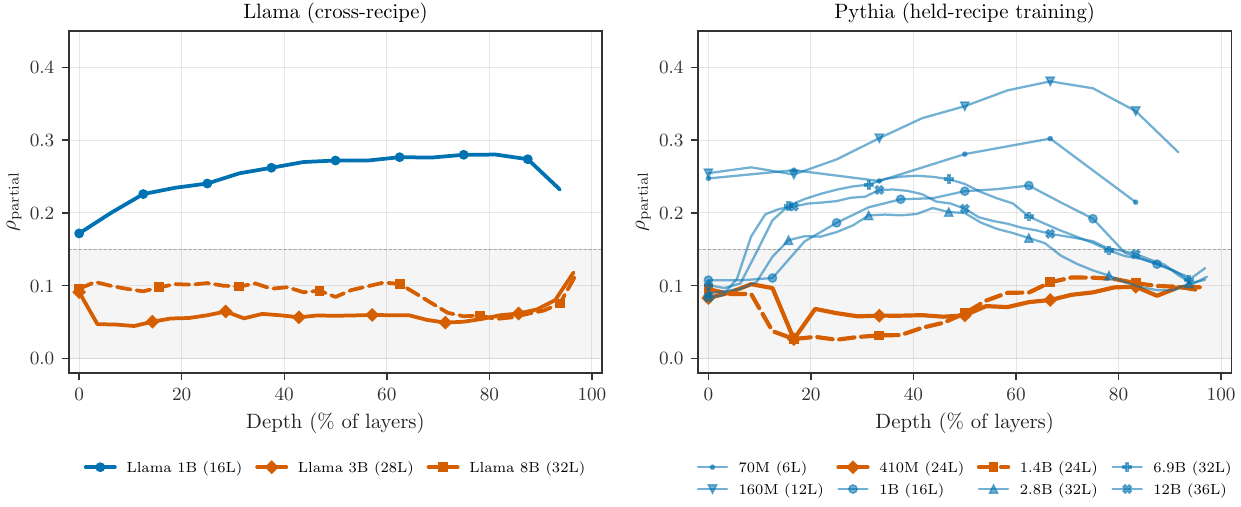}
\caption{Within Pythia's controlled suite (right), configuration
  determines observability; the same within-family discontinuity
  appears observationally in Llama (left, cross-recipe). The y-axis
  is $\pcorr$, the partial Spearman
  correlation between a linear probe and per-token loss after
  controlling for max-softmax probability and activation norm, as
  defined in \cref{sec:method}. Left, Llama: 1B preserves the signal,
  while 3B and 8B do not. Right, Pythia controlled suite:
  configurations outside the 24-layer, 16-head class preserve the
  signal, while 410M and 1.4B at this configuration collapse across a
  $3.5\times$ parameter gap with hidden dimensions 1024 and 2048. The
  1.4B-deduped replication appears in \cref{tab:pythia_suite}. Same
  protocol, same token budget per hidden dimension. Collapse is
  discrete, not gradual.}
\label{fig:within_family_cliff}
\end{figure}

\cref{fig:within_family_cliff} previews the resulting within-family
cliffs under this confidence-controlled metric.

The deeper question is whether the surviving signal is
output-independent. We therefore introduce an output-controlled
residual, $\ocresid$, which additionally controls for what a trained
predictor on the final-layer representation can recover. Positive
$\ocresid$ means the mid-layer observer reads decision-quality
information not recovered by the tested output-side readout. Across
26 tested models, $\ocresid$ tracks the confidence-controlled probe:
where observability collapses, the output-controlled residual
collapses with it.

Pythia supplies the controlled test for observability collapse. Three runs sharing the
(24-layer, 16-head) configuration class collapse to
$\pcorr \approx \pythiacollapseR$ despite $3.5\times$ variation in
parameters and two Pile variants, while six other configurations
produce $\pcorr$ between $\pythiahealthyminR$ and
$\pythiahealthymaxR$. Because the
output-controlled residual collapses with observability, what is lost
is the output-independent component isolated by the protocol.
Checkpoint dynamics show a loss-monitorability dissociation. Both
matched-width configurations form the signal at the earliest measured
checkpoint and improve in perplexity. Training erases observability
in the (24-layer, 16-head) class while the healthy configuration
recovers. Within this held recipe, architecture does not prevent the
signal from appearing. It determines whether training preserves or
erases it. Signal engineering (\cref{def:signal_engineering})
makes signal preservation or amplification a training objective
alongside loss.

Cross-family evidence extends the pattern observationally. At matched
3B scale, \qwen{} and Llama differ by $\crossfamilygap\times$ with
non-overlapping probe-seed distributions, while Mistral~7B v0.3 and
Llama~3.1 8B share the same high-level 32-layer, 32-head, 4096-hidden shape
yet diverge. These comparisons are observational: within a held recipe,
configuration determines preservation; across recipes, recipe
choice shifts where collapse occurs. A WikiText-trained
observer transfers to downstream QA without task-specific training,
catching about one in eight errors missed by output-confidence
monitoring at a 20\% flag rate.

\cref{tab:evidence_hierarchy} summarizes the evidence hierarchy:
measurement hardening, controlled within-recipe evidence,
observational cross-recipe extension, and scoped downstream utility.
Each tier points to the same upstream property. Architecture
selection is a monitoring decision.

\begin{table}[!ht]
\centering
\small
\setlength{\tabcolsep}{4pt}
\caption{Evidence hierarchy. The first three rows establish the
  measurement as strong and hardened. The remaining rows separate
  controlled within-recipe evidence from observational cross-recipe
  extensions and scoped downstream utility.}
\label{tab:evidence_hierarchy}
\begin{tabular}{>{\raggedright\arraybackslash}p{0.34\textwidth}>{\raggedright\arraybackslash}p{0.40\textwidth}>{\raggedright\arraybackslash}p{0.20\textwidth}}
\toprule
Claim & Evidence & Evidence type \\
\midrule
Raw probes are confidence-contaminated & \nabsorbmodels{}-model confidence-control analysis & Cross-model \\
\addlinespace[4pt]
Confidence-independent signal exists & 20-seed hardening, shuffle/random tests & Cross-seed + null \\
\addlinespace[4pt]
Signal is output-independent & $\ocresid$, output-side MLP width sweep & Multi-method \\
\addlinespace[4pt]
Configuration determines observability & Pythia controlled suite (3 replications) & Controlled within recipe \\
\addlinespace[4pt]
Collapse is training-emergent & Pythia~1B vs.\ 1.4B checkpoint trajectories & Controlled within recipe \\
\addlinespace[4pt]
Phenomenon recurs across recipes & Llama/Qwen/Mistral/Phi/Gemma & Observational \\
\addlinespace[4pt]
Recipes shift which configurations collapse & Mistral~7B vs.\ Llama~3.1 8B at matched 32L/32H/4096d shape & Observational \\
\addlinespace[4pt]
Operational catch on downstream tasks & LM/SQuAD/MedQA/TruthfulQA & Scoped utility \\
\bottomrule
\end{tabular}
\end{table}
\FloatBarrier

\section{Method}
\label{sec:method}

Standard probe evaluations mix decision quality with output
confidence. Our method isolates the confidence residual: a linear
probe on frozen transformer activations predicts a
confidence-residualized binary target. All transformer weights are frozen
throughout. Only the linear observer and the output-side MLP control
are trained.

\medskip
\begin{samepage}
\begin{definition}[Observability]
\label{def:observability}
The \emph{observability} of a transformer at layer $l$, equivalently
its \emph{linear readability}, is the degree to which a linear probe
on $\resid{l}$ predicts per-token loss beyond what max-softmax
probability and activation norm explain.
The scalar estimator is $\pcorr$ (partial Spearman correlation
controlling for these two covariates); the output-controlled residual
statistic, $\ocresid$, is the partial Spearman correlation obtained
after additionally controlling for a trained predictor on final-layer
activations. ``Decision quality'' refers to this
confidence-residual loss signal, not to a broader notion of reasoning
correctness.
\end{definition}
\end{samepage}
\medskip

\begin{samepage}
\begin{definition}[Observability collapse]
\label{def:collapse}
An architectural configuration undergoes \emph{observability
collapse} under a training recipe when $\pcorr$ falls to the
detection floor (\cref{sec:signal}) and the output-controlled
residual $\ocresid$ approaches zero, so that linearly readable,
output-independent decision quality is not preserved at any
measured layer. This is a collapse of monitorable signal, not of
predictive capability. Checkpoint trajectories distinguish
training-time erasure from static non-formation
(\cref{sec:pythia}).
\end{definition}
\end{samepage}
\medskip

\begin{samepage}
\begin{definition}[Signal engineering]
\label{def:signal_engineering}
\emph{Signal engineering} is the design of training procedures
that preserve or amplify specified target signals as readable from
activations of the frozen trained model. Whereas activation
monitoring takes a frozen model and asks what can be read, signal
engineering takes a target signal and asks what training preserves
or amplifies it.
\end{definition}
\end{samepage}
\medskip

Observability measures what a linear readout can extract, not what
the model's own computation uses. We use ``monitorability'' as the
broader operational property monitors depend on; ``observability''
is the linearly-readable instance this paper measures. The
confidence covariates test whether the mid-layer signal survives
output confidence at generation time; the output-side readout tests
whether it survives what a trained final-layer predictor can recover.

\medskip
\textbf{Scope and assumptions.}
We study frozen autoregressive transformers, per-token cross-entropy
loss, max-softmax and activation-norm covariates, linear mid-layer
observers, and output-side MLP readouts. This is a deliberately
narrow definition of decision-quality observability, not a claim
about reasoning correctness or semantic truth. The restriction is the
point. If a weak, single-pass linear observer reads a signal that
output-side predictors miss, the model preserved monitorable
information. If even nonlinear probes and full layer sweeps do not
recover it, the architecture-recipe did not preserve that signal under
this protocol.

\medskip
\textbf{Target construction.}
For each non-padding token position $i$ in the evaluation corpus, we
compute per-token cross-entropy loss $\ell_i$, max-softmax probability
$\hat{p}_{\max,i}$, and $\lVert \mathbf{h}_i \rVert$, the L2 norm of
the residual stream at layer $l$ (block output; equivalently the input
to layer $l+1$). An ordinary least squares (OLS) regression $\ell_i \sim \hat{p}_{\max,i} +
\lVert \mathbf{h}_i \rVert$ fitted on the training split produces
residuals $\epsilon_i = \ell_i - \hat{\ell}_i$. The same fitted
coefficients are applied without refitting to validation and test
tokens, so the residual target on held-out data uses train-fit
residualizer coefficients. The binary target is:
\begin{equation}
  y_{\mathrm{resid},i} = \mathbf{1}[\epsilon_i > 0].
  \label{eq:target}
\end{equation}
The target is approximately balanced: OLS residuals are mean-zero by
construction and empirically near-symmetric across WikiText-103
tokens. Binary cross-entropy supervision rather than regression on
the residual magnitude is a design choice analyzed in
\cref{sec:signal}. Because $\lVert \mathbf{h}_i \rVert$ is
layer-specific, the binary target is constructed separately for each
model and layer. Layer profiles therefore measure the readable
confidence residual signal at each depth under that layer's own norm
control.

\medskip
\textbf{Observer probe.}
The observer is a linear head on frozen residual stream activations at
layer $l$:
\begin{equation}
  o(\resid{l}) = \mathbf{w}^\top \resid{l} + b.
  \label{eq:observer}
\end{equation}
Training uses binary cross-entropy on $\btarget$ with Adam
(lr $= 10^{-3}$, batch size 4096, weight decay $10^{-4}$) for 20
epochs. The probe has $d + 1$ parameters. No activation function, no
hidden layer.

\medskip
\textbf{Evaluation: partial correlation.}
All reported correlations are Spearman rank partial correlations
controlling for confounds $\mathbf{c} = [\conf, \anorm]$. After
rank-transforming all variables, we project out the covariates:
\begin{equation}
\begin{aligned}
  \pcorr(s_\theta, \loss \mid \mathbf{c})
  &= \mathrm{corr}\!\bigl(
      r_{s_\theta \mid \mathbf{c}},\;
      r_{\loss \mid \mathbf{c}}
    \bigr) \\
  r_{x \mid \mathbf{c}}
  &= x - \mathbf{C}(\mathbf{C}^\top \mathbf{C})^{-1}\mathbf{C}^\top x.
\end{aligned}
\label{eq:pcorr}
\end{equation}
Here $s_\theta = o(\resid{l})$ is the probe score and $\mathbf{C}$ is the matrix of
ranked covariates with an intercept column. Spearman
correlation is invariant to monotone transformations of a scalar
score. Logit-level temperature scaling can reorder max-softmax values
across examples in multiclass settings, so the nonlinear MLP and
logit-entropy checks in \cref{sec:signal} serve as the empirical
robustness check.

\medskip
\label{sec:output_control}
\textbf{Output-controlled residual.}
An MLP ($d \to 64 \to 1$, ReLU) trained on last-layer activations to
predict loss $\hat{\loss}_\psi^{(L)}$ is the output-side control. The
MLP is trained with Adam (lr $= 10^{-3}$, batch size 1024, weight
decay $10^{-4}$) for 20 epochs on the probe's training split; its
predictions on the held-out evaluation tokens form
$\hat{\loss}_\psi^{(L)}$:
\begin{equation}
  \ocresid = \pcorr\!\left(
    s_\theta^{(l)},\; \loss \mid \mathbf{c},\; \hat{\loss}_\psi^{(L)}
  \right).
  \label{eq:ocresid}
\end{equation}
Positive $\ocresid$ means the mid-layer observer reads information
not recovered by the trained final-layer predictor. A width sweep (64--512
units) tests whether this result depends on predictor capacity
(\cref{sec:signal}). \cref{tab:monitoring_comparison} maps each
monitoring signal to the questions it can answer.

Each level of this hierarchy tests against a different alternative
explanation. Confidence control tests whether the probe merely
recapitulates output confidence. Output control tests whether the
mid-layer signal is merely re-reading information recoverable from
the final-layer representation by the tested output-side predictors.
The observer is complementary to those predictors when $\ocresid$ is
positive; when $\ocresid$ collapses together with $\pcorr$, this
complementary, output-independent component is no longer preserved
in linearly readable form under this protocol. Under a held recipe,
this is what it means for architecture to determine observability: it
determines whether the output-independent component that activation
monitors need survives. We use
$\pcorr \leq \detectionfloor$ as the empirical detection floor because random,
shuffled, and underpowered probes cluster near or below this range
(\cref{sec:signal}). Within the Pythia controlled suite, healthy
configurations begin at $\pcorr \geq \pythiahealthyminR$ (exact
minimum $0.208$, Pythia~2.8B), with the gap between these thresholds
treated as an indeterminate margin within that suite.

\begin{table}[t]
\centering
\small
\caption{What each monitoring signal establishes. The
  output-controlled residual $\ocresid$ is the only single-signal
  test of output-independence.}
\label{tab:monitoring_comparison}
\begin{tabular}{@{}lcccc@{}}
\toprule
Signal / protocol
  & \makecell{Predicts\\loss}
  & \makecell{Removes\\confidence}
  & \makecell{Tests output\\independence}
  & \makecell{Detects\\collapse} \\
\midrule
Output confidence              & \checkmark &            &            &            \\
Raw activation probe           & \checkmark &            &            & Weakly     \\
$\pcorr$ (conf.-controlled)    & \checkmark & \checkmark &            & Partial    \\
$\ocresid$ (output-controlled) & \checkmark & \checkmark & \checkmark & \checkmark \\
Nonlinear / layer sweeps       & \checkmark & \checkmark & Partly     & Artifact check \\
Pythia controlled suite        & \checkmark & \checkmark & \checkmark & Controlled (within recipe) \\
\bottomrule
\end{tabular}
\end{table}

\medskip
\textbf{Seed agreement.}
Seed agreement $\sagree$ is the mean pairwise Spearman correlation of
observer scores across $k$ independently initialized probes on the same
frozen activations:
\begin{equation}
  \sagree = \frac{2}{k(k-1)} \sum_{i<j} \rho_s(o_i, o_j).
  \label{eq:sagree}
\end{equation}
Observed $\sagree$ on frozen activations typically exceeds $0.9$,
consistent with convergence to a single direction across
initializations. Full seed-count analysis in \cref{sec:signal}.

\medskip
\textbf{Target validity.}
We construct the target by residualizing against confidence, then
evaluate against the same covariates. Six empirical tests address this
circularity, organized as one construction test, three selectivity
tests, and two null-distribution tests (full results in
\cref{sec:signal}). The construction test confirms that residualizer
fitting does not leak: a separate residualizer-fit split preserves
the regime on four representative models. Selectivity tests confirm
that the target is not trivially recoverable: hand-designed
activation statistics fail under the same target, the probe fails on
Llama~3B under the identical target, and training the probe on C4
web text under the same target-construction protocol fails. Null-distribution tests confirm that the signal is not a
probing artifact: random untrained probes and a shuffle test on
randomized labels both produce near-zero $\pcorr$.

\medskip
\textbf{Evaluation data.}
Core observability probes train and evaluate on
WikiText-103~\citep{merity2017pointer} with standard splits.
Probes train on the train split. Layer selection is two-stage: a
seed-42 sweep produces a candidate set (the top four layers by
seed-42 $\pcorr$, depth-filtered to $\le 80\%$ of total depth,
plus the seed-42 argmax), and the reported layer is the argmax of
the 7-seed mean (seeds 43--49) over candidates. Reported $\pcorr$
is the 7-seed distribution at that layer; the top-three layer
flatness in \cref{sec:appendix_validation} bounds
selection-induced optimism. The
standard token budget is 350 examples per hidden dimension (ex/dim),
where an example is one non-padding token position.
\cref{sec:signal} reports 20-seed hardening on GPT-2~124M,
\cref{sec:architecture} reports frozen-probe transfer to downstream
tasks, \cref{sec:appendix_cross_domain} reports cross-domain
transfer to C4 and other corpora, and
\cref{sec:appendix_validation} reports test-split confirmation and
layer-selection sensitivity.
Because the observer is a linear direction in a $d$-dimensional
residual stream, cross-scale comparisons require token budgets scaled
by hidden dimension. On \qwen{}~0.5B ($d = 896$), a seven-point sweep
from 150 to 1{,}000 ex/dim reveals a budget threshold between 450
and 600 ex/dim: below 450, $\pcorr$ is precise but near the detection
floor;
above 600, the signal rises to the reported healthy value
(\cref{sec:appendix_token_budget}). Low-budget estimates have low
seed variance, so small confidence intervals do not guarantee
adequate power. All reported cross-scale comparisons use budgets
above each model's empirically adequate threshold. Collapsed models
(Llama~3B, 8B; Pythia~410M, 1.4B) remain collapsed at absolute token
budgets well above the detection regime, so underpowering does not
explain collapse.

\medskip
\textbf{Statistical model.}
Cross-family comparisons use a mixed-effects model with per-seed
observations, random intercepts per model (absorbing within-model
correlation from shared data and layer selection), and fixed effects
for $\log_{10}(\text{params})$ and family:
\begin{equation}
  \pcorr{}_{\,ij} \sim \log_{10}(N_i) + \mathrm{family}_i + (1 \mid \mathrm{model}_i).
  \label{eq:mixed}
\end{equation}
Full specification and variance decomposition in
\cref{sec:architecture}. Family-level significance is assessed via a
Monte Carlo permutation test (50{,}000 samples) on model-mean $\pcorr$,
with $\log_{10}(\text{params})$ residualized before permutation
(\cref{sec:architecture_stats}). The protocol isolates per-token
decision quality from output confidence: the residual signal it
measures is what activations expose beyond the tested output-side
readouts.

\section{Isolating a monitorable decision-quality signal}
\label{sec:signal}

Before observability can serve as an architecture-and-training
design axis, the signal it measures must be separated from output
confidence, target-construction artifacts, and information already
recoverable by tested output-side readouts. This section isolates
that signal on WikiText-103, using
GPT-2~\citep{radford2019language} as the hardening case and
cross-model tests where needed. The goal is not measurement for its
own sake: it is to identify a monitorable decision-quality component
whose preservation can then be compared across architectures and
training recipes.

\medskip
\textbf{Raw probes are confidence-contaminated.}
On GPT-2~124M, raw Spearman correlation between a linear probe and
per-token loss is $\rawspearman$, but only $\stdcontrol$ survives
after controlling for max-softmax and activation norm. Across
\nabsorbmodels{} models in \nfamilies{} families, controlling for
confidence removes $\confabsorbmean\% \pm \confabsorbstd\%$ of the
raw signal,
ranging $\confabsorbmin\%$--$\confabsorbmax\%$ across the cohort
(per-model breakdown in \cref{sec:appendix_confidence}). On this
decision-quality target, half to three-quarters of what probes find
in hidden states is output confidence in disguise.

\begin{figure}[!t]
\centering
\includegraphics[width=0.80\textwidth,alt={Bar chart showing
  correlation dropping from 0.514 raw Spearman to 0.289 after max-softmax
  control, 0.289 after softmax and norm, 0.211 after adding logit
  entropy, and 0.319 after nonlinear MLP
  control.}]{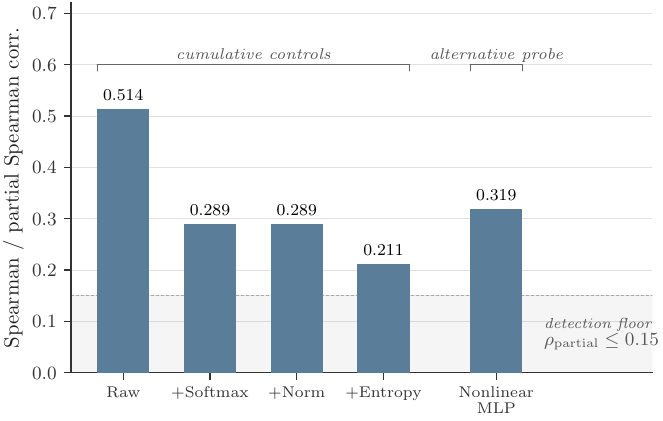}
\caption{Control sensitivity on GPT-2~124M. The first four bars are
  cumulative: raw Spearman ($\rawspearman$), then adding max-softmax
  control ($\softmaxonly$), then norm ($\stdcontrol$), then logit
  entropy ($\entropycontrol$). The fifth bar is an independent
  nonlinear MLP control ($\nonlincontrol$), confirming the residual
  is not a nonlinear function of max-softmax confidence and
  activation norm.}
\label{fig:waterfall}
\end{figure}

Two additional robustness checks do not eliminate the residual
signal (\cref{fig:waterfall}). A nonlinear MLP trained on
$[\conf, \anorm]$ to predict loss does not reduce the residual
signal: $\pcorr = \nonlincontrol$, compared with $\stdcontrol$ under
the linear standard control.
The nonlinear MLP can represent Platt scaling and other scalar
recalibrations of max-softmax confidence as special cases, so the
signal is not explained by miscalibrated confidence. Logit entropy
as a third covariate absorbs a further $\entropyabsorbpct\%$
($\pcorr = \entropycontrol$). The observer partially reads the
shape of the output distribution.

Raw correlations are easy to produce. Survival after controlling
for confidence is the bar.

\medskip
\textbf{A learned linear observer recovers a stable residual signal.}
Every hand-designed activation statistic loses signal after
controlling for confidence. On GPT-2~124M, Forward-Forward goodness, active ratio,
activation entropy, and activation norm all produce near-zero or
negative $\pcorr$ after controlling for confidence and activation
norm (\cref{tab:hand_designed_baselines}).

\begin{table}[t]
\centering
\caption{Hand-designed activation statistics on GPT-2~124M (layer~11).
All collapse to near-zero partial correlation under confidence controls.}
\label{tab:hand_designed_baselines}
\begin{tabular}{lc}
\toprule
Observer & Partial corr \\
\midrule
FF goodness ($\sum h_i^2$)  & $-0.010$ \\
Active ratio                & $-0.057$ \\
Activation entropy          & $-0.110$ \\
Activation norm             & $-0.002$ \\
\midrule
Learned linear binary       & $\mathbf{+0.282}$ \\
\bottomrule
\end{tabular}
\end{table}

A linear probe trained on the residualized binary target
(\cref{eq:target}) recovers $\pcorr = \hardpcorr \pm \hardstd$ on
frozen GPT-2~124M (20 seeds at layer~11, the hardening reference
layer; \cref{eq:observer,eq:pcorr}) with seed agreement $\hardsagree$
(\cref{eq:sagree}). The standard 7-seed protocol at the peak layer~8
yields $\corepcorr$. Random untrained probes produce near-zero
$\pcorr$ on tested architectures
(\cref{tab:cross_family_scaling}). Binary supervision produces higher
seed agreement than regression because the heads converge on a shared
decision boundary (\cref{sec:appendix_confidence}).

\medskip
\textbf{Validity checks address target and probe artifacts.}
The target construction residualizes against confidence and the
evaluation applies the same covariates (\cref{sec:method}). Six tests
in three categories address the resulting circularity: a construction
test, three selectivity tests, and two null-distribution tests.

The construction test confirms that residualizer fitting does not
leak. Fitting the OLS residualizer on a disjoint document pool and
applying those coefficients without refit to the probe-training and
evaluation pools preserves the healthy/collapsed regimes on
\residSplitNmodels{} representative models, with split-fit $\pcorr$
differing from the same-pool baseline by at most
$\residSplitMaxDelta$ (\cref{sec:appendix_residualizer_split}).
Selectivity tests confirm that the target is not trivially
recoverable: hand-designed activation statistics fail (above), the
observer fails on Llama~3.2 3B at $\llamapcorr$ under the identical
target, and training the observer on C4 web text under the same
target-construction protocol fails on \qwen{}~7B at
$\pcorr = \qSevenCfourWithin$ (\cref{sec:appendix_cross_domain}).
Null-distribution tests confirm that the signal is not a probing
artifact: random untrained probes produce near-zero $\pcorr$ across
tested architectures, and a probe trained on shuffled labels yields
$\pcorr = \shufflegptmean \pm \shufflegptstd$ across 10 permutations
on GPT-2~124M, against a real probe score of $\shufflegptreal$ that
exceeds every shuffled permutation.

\medskip
\textbf{The residual is partly output-independent.}
Observability exists at every GPT-2~124M layer, peaking at layer~8
of 12 ($\pcorr = \corepcorr$). The output-side control absorbs about
half of the layer~8 signal, but $\ocresid = \coreoc \pm \coreocstd$
survives (3 seeds; \cref{eq:ocresid}). Under this output-side
control, the observer is not merely re-reading final-layer
information from an earlier layer. It reads something the tested
output-side predictors do not recover.

A width sweep (64--512 units) on the output-side MLP applied to
\qwen{}~7B leaves the output-controlled residual stable at every width: capacity is
not the bottleneck. A 30-resample document-level bootstrap on
\qwen{}~7B yields $\pcorr = \bootmean$ (95\% CI $[\bootlo, \boothi]$),
stable under data resampling. A matched two-layer MLP probe is
statistically equivalent to the linear probe across \tostNmodels{}
models (TOST at margin $\pm \tostMargin$; \cref{sec:appendix_tost}).
In Llama~3.2 3B, a swept-HP nonlinear probe with held-out
hyperparameter selection scores $\llamaThreeNLmlp$, well below the
healthy floor: a weak linear readout does not explain the later
collapse results. The observer direction lies in a low-variance
subspace that variance-dominant and reconstruction-trained
decompositions partially obscure (\cref{sec:appendix_geometry}).

\cref{sec:pythia} identifies under controlled training the
architectures that preserve this output-independent component and
those that erase it.

\section{Pythia: A Controlled Within-Recipe Test}
\label{sec:pythia}

Pythia supplies the controlled architecture test. Holding corpus,
tokenizer, optimizer family, and schedule template fixed, the
eight-size suite varies architecture across scale. Under this held recipe, every
tested run sharing the 24-layer, 16-head configuration class
collapses to $\pcorr \approx \pythiacollapseR$, while six other
configurations occupy a separated healthy band
($\pythiahealthyminR$ to $\pythiahealthymaxR$). We refer to this
discrete loss of readable decision-quality signal as \emph{observability
collapse}. This section establishes the controlled separation, rules
out layer-selection, readout-capacity, and measurement-artifact
alternatives, and shows that collapse is training-emergent rather than
a static representational incapacity.

The Pythia suite~\citep{biderman2023pythia} holds corpus, tokenizer,
optimizer family, and schedule template constant across eight sizes
from 70M to 12B. Learning rate and batch size scale with model size
while architecture varies. The three 24-layer, 16-head
collapse points are Pythia~410M ($\pythiaFourTenpcorr$), Pythia~1.4B
($\pythiaOneFourpcorr$), and Pythia~1.4B-deduped
($\pythiaOneFourDeduppcorr$, same architecture as 1.4B but trained
on the deduplicated Pile~\citep{gao2021pile} variant), each measured
across 7 probe
seeds. The three fall within $\pythiatriwidth$ of each other
despite $3.5\times$ variation in parameters, $2\times$ in hidden
dimension, $2\times$ in head dimension, and two Pile variants. At
410M the seed variance is $\pm \pythiaFourTenstd$, the tightest
measurement in the paper.

Six other Pythia configurations, spanning depths 6--36 and head
counts 8--40, produce $\pcorr$ between $\pythiahealthyminR$ and
$\pythiahealthymaxR$. Pythia~1B (16-layer, 8-head, $\pythiaOnepcorr$) sits between the
two collapsed models in parameter count but produces healthy
observability, ruling out the 400M--1.5B size band as the collapse
trigger. The collapse is tied to the (24-layer, 16-head) configuration class,
not to scale.

\begin{table}[ht]
\centering
\caption{Pythia suite under held-recipe training (The Pile, same
  tokenizer, same optimizer family). Three runs sharing the (24 layers,
  16 heads) configuration class collapse (bold), replicated across a
  $3.5\times$ parameter gap (410M vs.\ 1.4B), $2\times$ hidden
  dimension and $2\times$ head dimension, and two Pile variants
  (regular vs.\ deduplicated at 1.4B). Six other configurations produce
  $\pcorr$ between $+0.21$ and $+0.38$. The output-controlled
  residual $\ocresid$ collapses at the same three runs
  (near zero or negative), while healthy configurations range $+0.09$ to $+0.17$.
  All values are 7-seed means on held-out seeds
  43--49 (layer selected via seed 42). Standard deviations are across
  seeds.}
\label{tab:pythia_suite}
\small
\begin{tabular}{lccccS[table-format=+1.3]S[table-format=1.3]S[table-format=+1.3]}
\toprule
Model & Layers & Heads & Hidden & Head\_dim & {$\pcorr$} & {Seed std} & {$\ocresid$} \\
\midrule
Pythia-70M           & 6  & 8  & 512  & 64  & 0.301 & 0.001 & 0.147 \\
Pythia-160M          & 12 & 12 & 768  & 64  & 0.382 & 0.004 & 0.169 \\
Pythia-410M          & 24 & 16 & 1024 & 64  & {$\mathbf{0.105}$} & {$\mathbf{0.001}$} & {$\mathbf{0.043}$} \\
Pythia-1B            & 16 & 8  & 2048 & 256 & 0.246 & 0.012 & 0.120 \\
Pythia-1.4B          & 24 & 16 & 2048 & 128 & {$\mathbf{0.106}$} & {$\mathbf{0.006}$} & {$\mathbf{0.003}$} \\
Pythia-1.4B-deduped  & 24 & 16 & 2048 & 128 & {$\mathbf{0.100}$} & {$\mathbf{0.007}$} & {$\mathbf{-0.012}$} \\
Pythia-2.8B          & 32 & 32 & 2560 & 80  & 0.208 & 0.007 & 0.088 \\
Pythia-6.9B          & 32 & 32 & 4096 & 128 & 0.240 & 0.012 & 0.110 \\
Pythia-12B           & 36 & 40 & 5120 & 128 & 0.238 & 0.005 & 0.097 \\
\bottomrule
\end{tabular}
\end{table}

The right panel of \cref{fig:within_family_cliff} shows layer-wise
profiles for all eight Pythia sizes: six configurations peak at
mid-to-late depth between $\pythiahealthyminR$ and
$\pythiahealthymaxR$, while the two 24-layer, 16-head
configurations (410M and 1.4B) stay flat near $\pythiacollapseR$
across every depth. The deduplicated 1.4B replication confirms
the same pattern (\cref{tab:pythia_suite}). At the final checkpoint, the collapsed configurations show no
mid-layer peak. The measured residual-stream states show no layer at
which healthy-range observability remains.
The left panel shows Llama under a different
training recipe, reproducing the same within-family discontinuity at
a different configuration. Llama~3.2 1B is healthy at (16-layer,
32-head), while 3.2 3B and 3.1 8B collapse at (28-layer, 24-head)
and (32-layer, 32-head).

\begin{figure}[t]
\centering
\includegraphics[width=\textwidth,alt={No layer choice rescues the
  Pythia (24 layers, 16 heads) collapse. Eight panels in a 2-by-4 grid,
  one per canonical Pythia size (70M, 160M, 410M, 1B, 1.4B, 2.8B,
  6.9B, 12B), show partial correlation across every layer. The two
  collapse configurations (410M and 1.4B, both 24 layers and 16
  heads) stay flat near 0.10 at every depth, rendered in vermillion.
  The six healthy configurations peak at mid-to-late depth between
  0.21 and 0.38, rendered in blue. A shared detection floor band
  below 0.15 and a dashed healthy-floor reference at 0.21 appear on
  every panel; each profile's peak layer is marked with a
  black-edged circle.}]{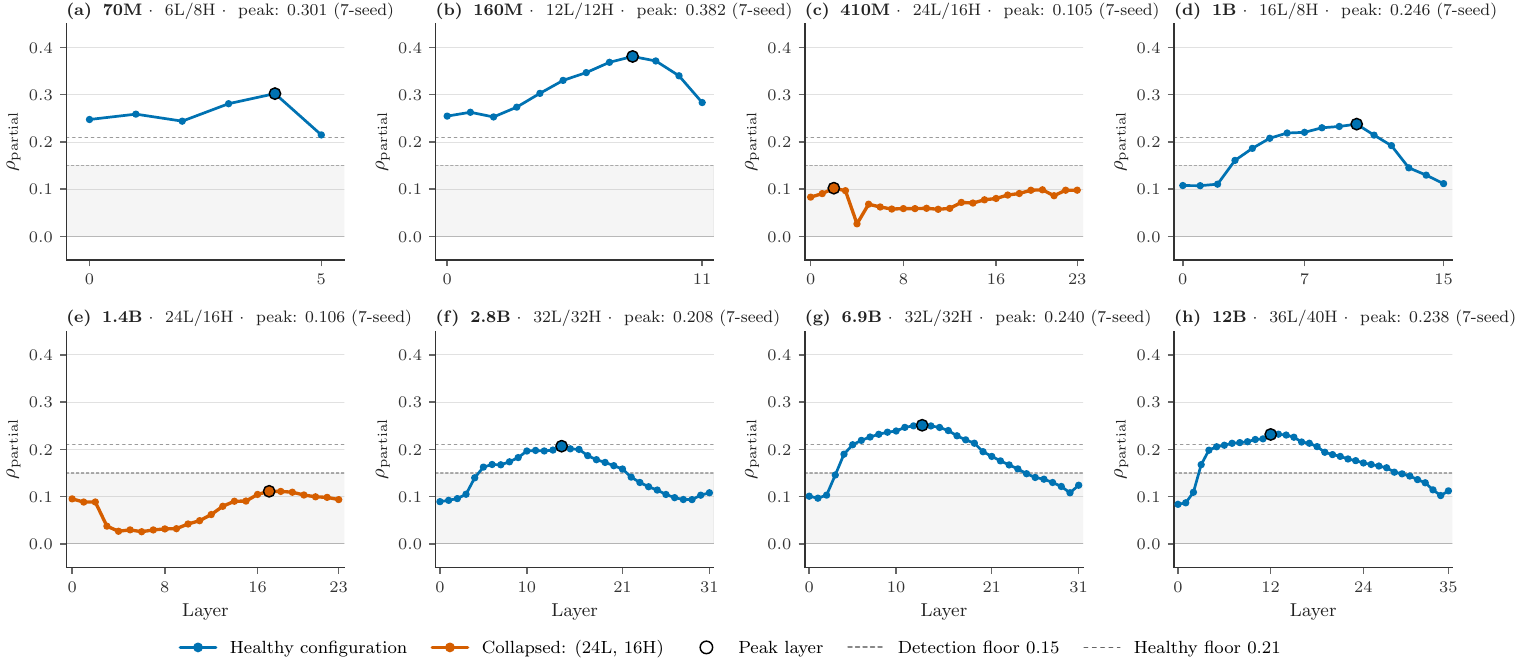}
\caption{Pythia layer profiles under controlled training. Six healthy
  configurations peak at mid-to-late depth. Two collapse
  configurations stay flat across all layers. No layer choice rescues
  the collapse.}
\label{fig:pythia_layers}
\end{figure}

\medskip
The three replications form three controlled comparisons, two
holding the (24-layer, 16-head) class constant while varying width or corpus,
and one holding the Pythia recipe constant while varying
configuration. Varying width, head dimension, and parameter count within
the same configuration class (Pythia~410M vs.\ 1.4B) does not
recover observability. Switching from regular to deduplicated Pile
at matched architecture (Pythia~1.4B vs.\ 1.4B-deduped) does not
recover it. Varying configuration at held training (other Pythia
sizes) produces the full range of observability values between
$\pythiahealthyminR$ and $\pythiahealthymaxR$
(\cref{fig:pythia_layers}). Two Pile variants, two sizes with a
$3.5\times$ parameter gap, and three 7-seed runs each produce
$\pcorr \approx \pythiacollapseR$.

\medskip
\textbf{Collapse is configuration-specific, not monotone in depth or heads.}
Collapse is not a monotonic function of any single architectural
axis. Deeper models do not collapse harder: 32-layer Pythia (2.8B at
$\pythiaTwoEightpcorr$, 6.9B at $\pythiaSixNinepcorr$) and 36-layer
Pythia (12B at $\pythiaTwelvepcorr$) produce healthy signal. Higher
head counts do not collapse harder: 32-head and 40-head
configurations are healthy. Depth and head count co-occur at both
collapse points (depth=24 AND heads=16) and no other Pythia size
matches either value. The two axes are perfectly confounded within
this suite.

\medskip
\textbf{Baselines and validity.}
Pythia~70M's random untrained probe produces $\pythiaSeventyrandom$,
elevated relative to larger Pythia sizes but still $\pythiaSeventyRatio\times$ below
the learned probe's $\pythiaSeventypcorr$. The finding holds with
this elevated baseline. The target validity tests from
\cref{sec:signal} apply unchanged here: the probe training protocol
is identical, residualizing against the same confidence covariates.

\medskip
\textbf{Within-Pythia statistical test.}
If you were allowed to label any two Pythia configurations as
collapsed, the actual (24-layer, 16-head) architecture label
produces the largest gap. An exact permutation test formalizes this:
under the null (arbitrary 2-vs-6 partitions), the observed gap of
$\pythiaobsgap$ is the unique maximum of the $\binom{8}{2} = 28$
possible partitions ($p = \pythiapermp$). Including the 1.4B-deduped
replication as a third collapsed point lowers the permutation
$p$-value to $\pythiapermpdedup$ ($1/84$). F-test, $\eta^2$, and
leave-one-configuration-out robustness appear in
\cref{sec:appendix_hardening}.

\medskip
\textbf{The collapse is signal, not probe failure.}
A shuffle test at the collapse configuration confirms the residual
value is real signal, not probe artifact. At Pythia~1.4B layer~17,
probes trained on randomly permuted labels yield
$\pcorr = \pythiaShufflemean \pm \pythiaShufflestd$ across 10
permutations, while the real probe scores $\pythiaOneFourpcorr$ and
exceeds every shuffled run. The permutation-null Gaussian
approximation, code-path sanity check, and GPT-2 cross-comparison
appear in \cref{sec:appendix_hardening}.

\medskip
\textbf{Linear probe is not the bottleneck.}
A two-layer MLP probe with held-out hyperparameter selection does
not close the gap at Pythia~410M or Pythia~1.4B
(\cref{tab:nonlinear_probe_absence}); the same conclusion at
Llama~3.2 3B is reported in \cref{sec:signal}. Across the three
collapse configurations under two training recipes, the
best tested nonlinear probe never crosses the healthy floor of
$\pythiahealthyminR$. The collapse reflects a representational
property rather than a limit of linear probing.

\begin{table}[ht]
\centering
\caption{Swept-hyperparameter nonlinear probe with held-out selection
  (fit on train, select HP on WikiText validation, report on WikiText
  test). Grid: hidden $\in \{64, 128\}$, lr $\in \{0.01, 0.001,
  0.0001\}$, epochs $\in \{20, 50\}$. $\Delta$ is swept-HP MLP minus
  linear on test. Above-floor configurations (healthy on the main
  validation protocol) gain $\le +0.07$ on test. The three collapse
  configurations, tested under two training recipes (Pythia
  held-recipe; Llama production) and at two layer choices for
  Llama~3B, gain $+0.014$ to $+0.043$ on test and remain below the
  validation healthy floor of \pythiahealthyminR. Test values may
  fall slightly below validation due to split sampling
  (\cref{sec:appendix_validation}); no tested probe recovers
  healthy-range signal where the linear probe fails.}
\label{tab:nonlinear_probe_absence}
\small
\begin{tabular}{llccc}
\toprule
Configuration & Peak layer & Linear test & Swept-HP MLP test & {$\Delta$} \\
\midrule
\multicolumn{5}{l}{\emph{Above-floor configurations (healthy on main validation protocol)}} \\
GPT-2~124M                       & L8  & $0.293$ & $0.339$ & $0.046$ \\
\qwen{}~0.5B                     & L19 & $0.219$ & $0.254$ & $0.035$ \\
\qwen{}~1.5B                     & L19 & $0.277$ & $0.325$ & $0.048$ \\
\qwen{}~3B                       & L25 & $0.286$ & $0.325$ & $0.038$ \\
\qwen{}~7B                       & L17 & $0.244$ & $0.295$ & $0.051$ \\
\qwen{}~14B                      & L30 & $0.228$ & $0.296$ & $0.067$ \\
Gemma~3~1B                       & L11 & $0.174$ & $0.198$ & $0.024$ \\
\midrule
\multicolumn{5}{l}{\emph{Collapsed configurations}} \\
Llama~3.2 3B                     & L0  & $0.088$ & $0.102$ & $0.014$ \\
Llama~3.2 3B (layer check)       & L27 & $0.104$ & $0.147$ & $0.043$ \\
Pythia~410M (24L, 16H)           & L2  & $0.126$ & $0.139$ & $0.014$ \\
Pythia~1.4B (24L, 16H)           & L17 & $0.129$ & $0.172$ & $0.043$ \\
\bottomrule
\multicolumn{5}{l}{\footnotesize Deltas computed from unrounded values.} \\
\end{tabular}
\end{table}

\FloatBarrier
\medskip
None of the tested alternative explanations accounts for the collapse. Signal could be confidence leakage, but
confidence covariates are applied throughout. It could be a
linear-readout limitation, but tested nonlinear probes do not cross
the healthy floor. It could be layer misselection, but collapsed
models show no healthy-range layer at any depth. It could be
underpowering, but collapsed models remain collapsed at token budgets
well above the budget threshold. It could be architectural
incapability, but both configurations form and strengthen the signal
early in training. What remains is whether the mid-layer signal is
merely redundant with the output representation. That is the test
$\ocresid$ performs.

\medskip
\textbf{Output-independence collapses at the same configurations.}
The output-controlled residual $\ocresid$ tracks $\pcorr$ across the
Pythia suite and collapses at the same three runs
(\cref{tab:pythia_suite}). Healthy configurations produce $\ocresid$
between $\pythiahealthyOClo$ (Pythia~2.8B) and $\pythiahealthyOChi$
(Pythia~160M). The three 24-layer, 16-head runs produce
$\pyFourTenOC$ (410M), $\pyOneFourOC$ (1.4B), and $\pyOneFourDedOC$
(1.4B-deduped):
near-zero or negative output-independent component at each collapse
point. Separation is wider on $\ocresid$ than on $\pcorr$ because
collapsed $\ocresid$ values reach zero or negative. At the
24-layer, 16-head class, the confidence-controlled signal falls to
the detection floor and the output-controlled component falls to
near zero. Architecture configuration therefore determines
output-independence under controlled training, not just partial
correlation. What collapses is not merely a metric but the
output-independent component the controls isolate.

The cross-family generalization of this property appears in
\cref{sec:architecture}: architectures that lose
confidence-controlled observability also lose output-independence
across the 26 models measured in the paper.

\medskip
\textbf{Observability collapse emerges during training.}
Both matched-width configurations form and strengthen the quality
signal early in training. We evaluate two Pythia configurations with
matched hidden dimension ($d = 2048$) at 10 checkpoints from
step~256 ($\approx$0.5B tokens) to step~143{,}000 ($\approx$300B
tokens), recomputing losses, confidence covariates, residual targets,
probe heads, and output-side readouts at each checkpoint under the
same protocol as the final-checkpoint experiments. At the earliest
measured checkpoint, both the 1B (16-layer, 8-head, healthy at
convergence) and the 1.4B (24-layer, 16-head, collapsed at
convergence) show healthy $\pcorr$ ($\pythiaOneEarliestpcorr$ and
$\pythiaOneFourEarliestpcorr$) and positive $\ocresid$
($\pythiaOneEarliestOC$ and $\pythiaOneFourEarliestOC$). By step~1{,}000 ($\approx$2B tokens),
the 1.4B configuration has strengthened further
($\pcorr = \pythiaOneFourMidpcorr$, $\ocresid = \pythiaOneFourMidOC$).
The 24-layer, 16-head architecture is capable of representing
the output-independent decision-quality signal early in training.

\begin{figure}[!t]
\centering
\includegraphics[width=0.63\textwidth,alt={Two-panel figure showing
  observability trajectories during Pythia training. Top panel:
  confidence-controlled partial correlation (rho partial) for Pythia
  1B (16 layers, 8 heads, blue) and Pythia 1.4B (24 layers, 16 heads,
  vermillion) across 10 checkpoints from 0.5B to 300B tokens seen.
  Both start in the healthy band above 0.21 at the earliest
  checkpoint. Both dip below the detection floor around step 16000.
  The 1B configuration recovers and converges near 0.25; the 1.4B
  configuration converges near 0.10. Bottom panel: output-controlled
  residual (r OC) over the same checkpoints. The 1B configuration
  ends at r OC 0.12; the 1.4B configuration ends near
  zero.}]{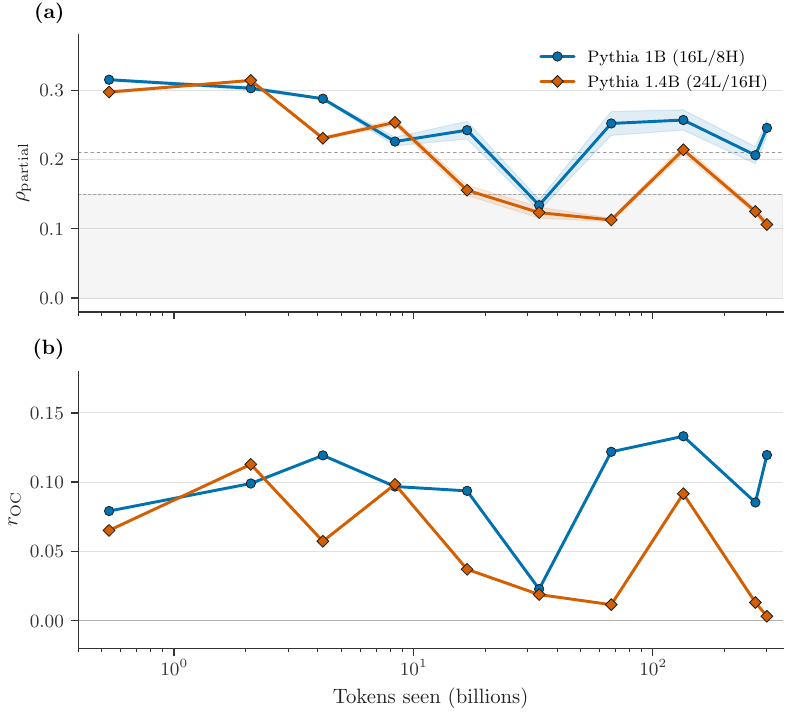}
\caption{Training improves prediction while erasing observability.
  Both configurations begin with healthy $\pcorr$ at the earliest
  measured checkpoint. Both dip during mid-training. The 16-layer,
  8-head configuration recovers and converges healthy. The 24-layer,
  16-head configuration converges collapsed. Final perplexities are
  comparable (18--21), so the divergence is monitorability, not
  predictive capability. Dashed lines in panel (a): detection floor
  (0.15) and healthy floor (0.21).}
\label{fig:checkpoint_dynamics}
\end{figure}

The trajectories diverge during mid-training
(\cref{fig:checkpoint_dynamics}). Both configurations dip below the
detection floor around step~16{,}000. The 1B configuration recovers
by step~32{,}000 and converges healthy
($\pcorr = \pythiaOnepcorr$, $\ocresid = \pythiaOneFinalOC$); the
1.4B configuration does not, converging after a transient recovery
at step~64{,}000 to $\pcorr = \pythiaOneFourpcorr$ and
$\ocresid = \pyOneFourOC$, matching the final-checkpoint collapse in
\cref{tab:pythia_suite}. The erasure is selective: the
output-independent fraction ($\ocresid / \pcorr$) drops from 36\%
at step~1{,}000 to 3\% at convergence in the 1.4B, while it grows
from 33\% to 49\% in the 1B (\cref{tab:checkpoint_dynamics}).

Perplexity declines monotonically in both configurations through
the observability dip. The 1B converges to perplexity
\pythiaOnePerplexity{} with $\pcorr = \pythiaOnepcorr$; the 1.4B
converges to lower perplexity \pythiaOneFourPerplexity{} with
$\pcorr = \pythiaOneFourpcorr$, so improved language-modeling loss
does not imply preserved observability. The healthy configuration
achieves both, so preservation does not require trading off
predictive performance. Predictive capability and observability are
not inherently coupled in either direction.

Architecture does not prevent the signal from appearing. It
determines whether training preserves or erases it. The collapse is
therefore not static incapacity: the signal forms, strengthens, and
is later erased. This is the empirical hinge for signal engineering
(\cref{def:signal_engineering}): an intervention window opens
between formation and erasure.

Under controlled training, configuration determines whether the
signal survives. \Cref{sec:architecture} asks whether independent
recipes reproduce the preservation-versus-collapse pattern at
different collapse points.

\FloatBarrier

\section{Across Independent Training Recipes}
\label{sec:architecture}

\Cref{sec:pythia} established the controlled within-recipe result:
under Pythia, configuration determines whether training preserves
observability. This section asks whether the
preservation-versus-collapse pattern recurs under independent
training recipes, and what ceiling it imposes on downstream
monitoring. Under a shared evaluation protocol with token budgets
scaled by hidden dimension, six families
(\qwen{}~\citep{yang2024qwen25},
Llama~\citep{grattafiori2024llama3},
Gemma~3~\citep{team2025gemma3}, Mistral~\citep{jiang2023mistral},
Phi-3~\citep{abdin2024phi3}, and GPT-2) show preserving, marginal,
and collapsed regimes. These comparisons are observational; Pythia
remains the controlled causal anchor.

\FloatBarrier

\medskip
\textbf{Two families preserve observability across scale.}
\qwen{} preserves observability across a $64\times$ parameter range:
$\pcorr$ stays between $\qwenrangelo$ and $\qwenrangehi$ from 0.5B
through 32B (\cref{tab:cross_family_scaling}), and the
output-controlled residual $\ocresid$ remains positive at every
scale tested. GPT-2 shows the
same scale-invariance under a different template: from 124M to
1.5B, $\pcorr$ remains between $\gptpcorrLo$ and $\gptpcorrHi$
(\cref{tab:gpt2_scaling}) and $\ocresid$ stays positive at every
size ($\gptOcMin$--$\gptOcMax$). Under fixed family templates,
scale alone does not change observability under these recipes.
Instruction tuning preserves the same pattern across tested Qwen
instruct sizes (\cref{sec:appendix_instruct}).

\begin{table}[t]
\centering
\caption{Observability across six architecture families under shared
evaluation protocol. Token budgets scaled by hidden dimension (350 ex/dim,
600 for \qwen{}~0.5B). Llama~1B matches high-observability families;
3B and 8B drop to near the detection floor.}
\label{tab:cross_family_scaling}
\small
\setlength{\tabcolsep}{4pt}
\begin{threeparttable}
\begin{tabular}{llccS[table-format=+1.3]S[table-format=1.3]S[table-format=+1.3]S[table-format=+1.3]S[table-format=+1.3]}
\toprule
Model & Family & Params & Peak layer & {$\pcorr$\tnote{a}} & {$\pm$ std} & {$\ocresid$} & {$\sagree$} & {Rand.\ head} \\
\midrule
\qwen{}~0.5B            & Qwen    & 0.5B  & L19 (79\%)    & 0.215 & 0.020 & 0.059 & 0.959 & -0.055 \\
Gemma~3~1B              & Gemma   & 1B    & L11 (42\%)    & 0.216 & 0.011 & 0.161 & 0.784 & -0.004 \\
Llama~3.2 1B            & Llama   & 1.2B  & L13 (81\%)    & 0.286 & 0.006 & 0.120 & 0.995 & 0.012 \\
\qwen{}~1.5B            & Qwen    & 1.5B  & L18 (64\%)    & 0.275 & 0.032 & 0.127 & 0.953 & -0.051 \\
GPT-2~XL                & GPT-2   & 1.5B  & L36 (75\%)    & 0.296 & 0.017 & 0.084 & 0.829 & 0.002 \\
\qwen{}~3B              & Qwen    & 3B    & L25 (69\%)    & 0.263 & 0.021 & 0.144 & 0.925 & 0.023 \\
Llama~3.2 3B            & Llama   & 3B    & L0 (0\%)      & 0.091 & 0.006 & 0.031 & 0.998 & -0.002 \\
Phi-3~Mini              & Phi     & 3.8B  & L17 (53\%)    & 0.300 & 0.002 & 0.144 & 0.958 & -0.008 \\
Gemma~3~4B              & Gemma   & 4.3B  & L6 (18\%)     & 0.191 & 0.007 & 0.059 & 0.802 & -0.010 \\
Mistral~7B              & Mistral & 7B    & L22 (69\%)    & 0.313 & 0.001 & 0.156 & 0.995 & 0.014 \\
\qwen{}~7B              & Qwen    & 7B    & L17 (61\%)    & 0.255 & 0.019 & 0.137 & 0.964 & -0.013 \\
Llama~3.1 8B            & Llama   & 8B    & L1 (3\%)      & 0.093 & 0.012 & -0.002 & 0.994 & 0.005 \\
\qwen{}~14B             & Qwen    & 14B   & L30 (62\%)    & 0.214 & 0.032 & 0.096 & 0.851 & 0.009 \\
\qwen{}~32B             & Qwen    & 32B   & L44 (69\%)    & 0.237 & 0.004 & 0.103 & 0.871 & -0.019 \\
\bottomrule
\end{tabular}
\begin{tablenotes}
\footnotesize
\item[a] Validation split (held-out seeds, $n = 7$ for all models). Across 12 cross-family models with test splits, rankings preserve and mean absolute gap is \testvalmeanabs\% (\cref{sec:appendix_validation}).
\item Llama~3B/8B peak at L0/L1: no layer exceeds $+0.12$; reported peak is argmax of a flat profile.
\end{tablenotes}
\end{threeparttable}
\end{table}

\medskip
\textbf{At matched 3B scale, Qwen and Llama diverge sharply.}
The difference is not subtle: \qwen{} retains
strong observability ($\qThreepcorr$) while Llama sits near the
detection floor ($\llamapcorr$), a $\crossfamilygap\times$ gap.
The per-probe-seed distributions do not overlap: every \qwen{} seed
($\qThreeseedlo$ to $\qThreeseedhi$) exceeds every Llama seed
($\llamaseedlo$ to $\llamaseedhi$). The gap survives a nonlinear
confidence check: under the nonlinear MLP on $[\conf, \anorm]$
(\cref{sec:signal}), Llama~3B falls to $\llamanonlin$.

\begin{figure}[!ht]
\centering
\includegraphics[width=0.90\textwidth,alt={Partial correlation vs parameter
  count for six model families. Qwen spans 0.21 to 0.28. GPT-2 spans 0.29
  to 0.30. Llama drops from 0.286 at 1B to 0.091 at 3B.}]{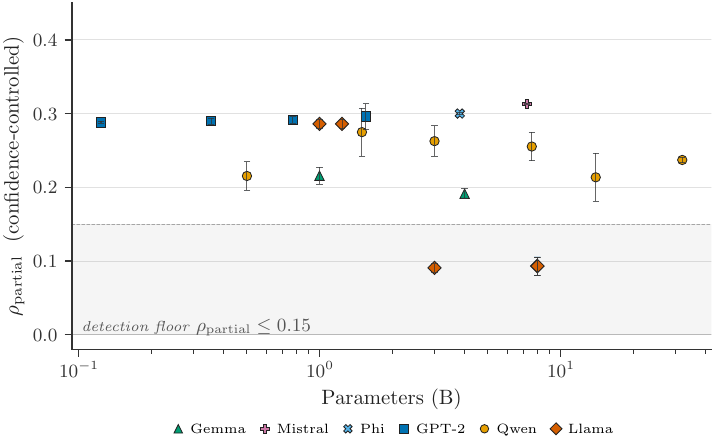}
\caption{Observability across six families. Llama (red) drops from
  $\llamaOnepcorr$ at 1B to $\llamapcorr$ at 3B. GPT-2 spans
  $\gptpcorrLo$ to $\gptpcorrHi$ across the four sizes. Non-Llama
  families remain above the detection floor. Shaded band: detection
  floor.}
\label{fig:cross_family}
\end{figure}

GPT-2~(124M--1.5B), Mistral~7B ($\mistralpcorr$), Phi-3~Mini
($\phipcorr$), Gemma~3~1B ($\gemmapcorr$), and Gemma~3~4B
($\gemmaFourpcorr$) confirm signal across four additional
families (\cref{fig:cross_family}).

\FloatBarrier
\medskip
\textbf{The Llama cliff.}
The Llama cliff is not gradual (\cref{fig:within_family_cliff},
left). Llama~3.2 1B produces healthy observability
($\pcorr = \llamaOnepcorr$), while Llama~3.2 3B falls near the
detection floor ($\llamapcorr$) and Llama~3.1 8B confirms the
collapsed regime ($\llamaEightpcorr$). The 1B-to-3B transition
changes both architecture (16-layer, 32-head, 2048-hidden vs.\
28-layer, 24-head, 3072-hidden) and distillation
provenance~\citep{llama32modelcard}, and the 3B-to-8B extension
crosses the 3.2/3.1 recipe boundary. The Llama comparison does not
isolate a single causal axis; Pythia supplies that controlled test.
It shows that the within-family discontinuity recurs under a
different recipe, at a different configuration from Pythia (Llama:
28-layer, 24-head; Pythia: 24-layer, 16-head).

\medskip
\textbf{Training recipe shifts which configurations collapse.}
Mistral~7B v0.3 and Llama~3.1 8B share the same high-level
32-layer, 32-head, 4096-hidden shape yet differ in observability
($\mistralpcorr$ vs.\ $\llamaEightpcorr$). Within a recipe,
configuration determines preservation; across recipes, recipe choice
shifts which configurations collapse. No tested
configuration collapses universally across recipes. Across recipes,
the architecture-recipe choice bounds the upstream ceiling on
observability. Signal engineering (\cref{def:signal_engineering})
makes recipe a design choice alongside architecture.

\begin{figure}[!ht]
\centering
\includegraphics[width=0.95\textwidth,alt={Output-independence and
  observability track together across 26 models; both vanish at the
  collapse configurations. Scatter plot of output-controlled residual
  (r subscript OC) against partial correlation (rho subscript
  partial). Healthy models from seven families cluster on a linear
  trend of slope 0.51 with a 95 percent bootstrap confidence band.
  Five collapsed runs (Llama 3B and 8B; Pythia 410M, 1.4B,
  and 1.4B-deduped) sit near the origin on both axes (rho partial
  near 0.10, r subscript OC near zero or negative).}]{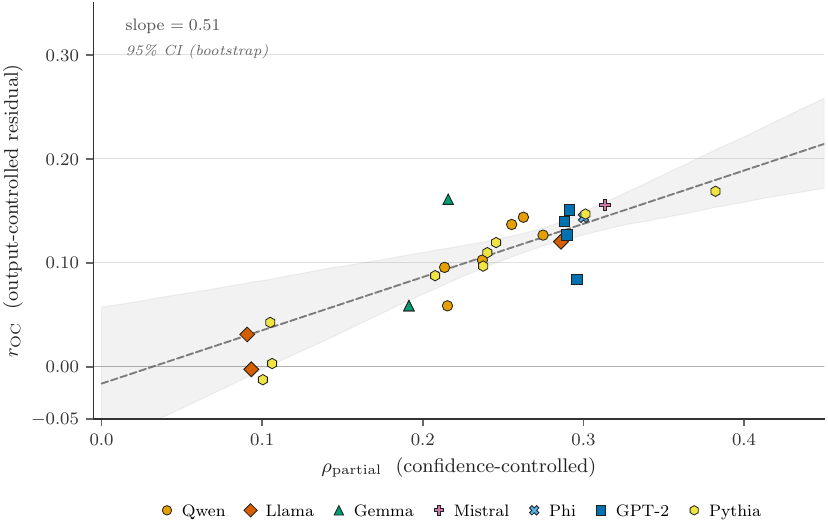}
\caption{Output-independence (against the tested final-layer readout) collapses with observability. Output-controlled residual
  ($\ocresid$) vs confidence-controlled partial correlation
  ($\pcorr$) across 26 models (17 cross-family plus 9 Pythia). Healthy
  configurations fall on a linear trend (dashed line, slope $\ocPcorrSlope$).
  Five collapsed runs (Llama 3B, Llama 8B, and three Pythia
  replications at the 24-layer, 16-head class) fall near the origin
  on both axes and are excluded from the healthy-configuration fit.
  Architecture-dependent observability and output-independence
  are two views of the same collapse phenomenon.}
\label{fig:oc_vs_pcorr}
\end{figure}

\FloatBarrier
\medskip
\textbf{Output-independence tracks the cross-family pattern.}
The output-controlled residual extends the within-Pythia result
across all measured families: models that lose
confidence-controlled observability also lose the output-independent
component measured by $\ocresid$ (\cref{fig:oc_vs_pcorr}).
Preserving configurations across all measured families fall on a
linear trend; five collapsed runs (Llama 3B, 8B; Pythia 410M, 1.4B,
1.4B-deduped) sit near the origin on both axes.

\medskip
\textbf{Statistical scope.}
\label{sec:architecture_stats}
The matched-scale 3B pair is the main cross-family comparison:
under identical protocol, evaluation data, and token budget per
hidden dimension, \qwen{}~3B and Llama~3B have non-overlapping
probe-seed distributions. Broader cross-family statistics are
sensitivity checks rather than the inferential anchor, because
four families contribute a single model. A permutation test on
model-mean $\pcorr$ ($p = \permp$) and a mixed-effects model
(\cref{eq:mixed}) both show family-level structure dominating
scale; full leave-one-family-out, Jonckheere-Terpstra,
analysis-of-covariance, and variance-decomposition analyses appear
in \cref{sec:appendix_hardening}. The controlled causal claim
remains the Pythia suite; the cross-family analyses show that
preservation-versus-collapse recurs under independent recipes.

\medskip
\textbf{Operational catch is bounded by error structure.}
Architecture-dependent observability sets an upstream ceiling, but
operational catch is also bounded by the error structure. A monitor
can only catch errors that are both present in the task and
separable from confidence by the observer score.
\cref{tab:flagging_cross_scale} therefore reports exclusive catch
across both preserving and collapsed configurations at fixed flag
rates. At 10\% flag rate, the observer catches confidence-invisible
language-modeling errors under this loss-based definition. For
language modeling, an error is a token with loss above the median,
so errors occupy half of all tokens. Exclusive catch is the fraction
of these errors flagged by the observer but not by confidence.

\begin{table}[t]
\centering
\caption{Exclusive error catches (observer finds, confidence misses) as a
percentage of all errors, across four architecture families at four flag
rates. The catch rate tends to be larger for higher-$\pcorr$ models at low flag rates but
converges to \catchsaturation\% at 20\%, indicating a ceiling set by the error
structure rather than by observability.}
\label{tab:flagging_cross_scale}
\begin{tabular}{llccccc}
\toprule
Model & Family & $\pcorr$ & 5\% & 10\% & 20\% & 30\% \\
\midrule
Mistral~7B        & Mistral & $0.313$ &   7.3\% &  11.4\% &  14.5\% &  12.6\% \\
GPT-2~124M        & GPT-2   & $0.288$ &   6.9\% &  10.8\% &  14.9\% &  16.5\% \\
\qwen{}~7B        & Qwen    & $0.255$ &   6.1\% &  9.8\% &  13.0\% &  12.7\% \\
\qwen{}~14B       & Qwen    & $0.214$ &   6.3\% &  10.4\% &  13.2\% &  12.8\% \\
Llama~3.2 3B      & Llama   & $0.091$ &   3.8\% &   8.4\% &  11.6\% &  12.0\% \\
Llama~3.1 8B      & Llama   & $0.093$ &   5.9\% &   9.7\% &  12.0\% &  11.4\% \\
\bottomrule
\end{tabular}

\smallskip\par\noindent\footnotesize Catch-rate columns are 3-seed means; $\pcorr$ is the 7-seed validation mean from \cref{tab:cross_family_scaling}. Flag rate is the fraction of tokens flagged by each monitor.
\end{table}

At low flag rates, higher observability tends to yield larger exclusive
catch. At 20\%, the models converge to a \catchsaturation\% band,
indicating a ceiling set by the structure of the error set rather
than by observability alone. Exclusive catch has a nonzero random baseline
because two rankers with fixed flag rates will disagree by
construction. On GPT-2~124M, an analytical random ranker
independent of confidence catches $\randbaselineten$\% at 10\% and
$\randbaselinetwenty$\% at 20\%. The observer exceeds random by
$\obsvsrandten$ percentage points at 10\% and $\obsvsrandtwenty$ at
20\%. The margin narrows as flag rate rises.

\medskip
\textbf{Downstream transfer near the language-modeling ceiling.}
The WikiText observer transfers across three question answering
(QA) tasks (SQuAD, MedQA, TruthfulQA), catching $\downstreammin$--$\downstreammax\%$
of confidence-invisible errors (errors flagged by the observer but
missed by output-confidence monitoring) at 20\% flag rate in seven of
nine model-task cells (\cref{sec:appendix_downstream}). On the
fluent-falsehood subset of TruthfulQA the observer is near chance
(area under the ROC curve, AUC $\tqaAUCqwen$, $\tqaAUCmistral$,
$\tqaAUCphi$ across three production instruct models), marking a method boundary supported
by consistency across three models:
token-level decision-quality observability does not reach locally
plausible semantic falsehoods. The WikiText-trained probe also transfers to C4 web text across five
families (\cref{sec:appendix_cross_domain}); the trained observer
generalizes across domains, while direct C4 training often fails,
indicating that clean calibration data matters. These rates establish
non-overlapping coverage relative to confidence rather than a
demonstrated margin over task-specific random rankers; the
language-modeling random-ranker analog is reported above, with
task-specific computation deferred to future work.

\begin{figure}[t]
\centering
\includegraphics[width=\textwidth,alt={Exclusive catches across 3
  production instruct models and 3 downstream tasks. Two-panel bar
  chart: left at 10 percent flag rate is variable by task and model;
  right at 20 percent flag rate shows 7 of 9 cells landing inside the
  10.9 to 13.4 percent downstream in-band range, close to the
  language-modeling ceiling: all Qwen 2.5 7B Instruct and Mistral 7B
  Instruct cells across SQuAD 2.0, MedQA-USMLE, and TruthfulQA, plus
  Phi-3 Mini Instruct on TruthfulQA. The 2 exceptions are Phi-3 Mini
  Instruct on SQuAD at 6.6 percent and on MedQA at 4.1
  percent.}]{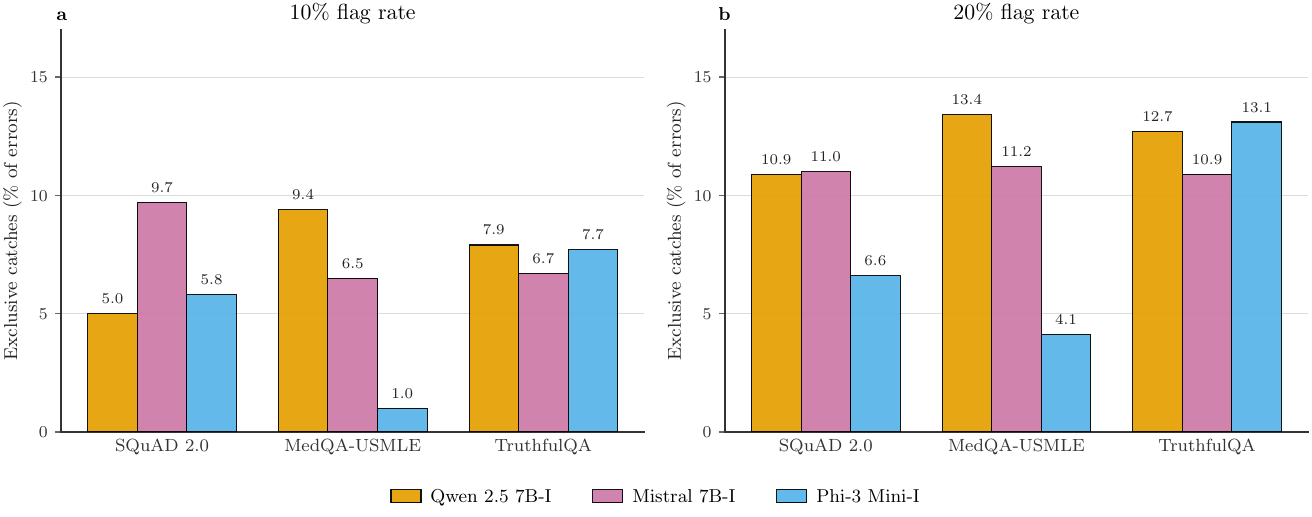}
\caption{Exclusive catch rates across three tasks and three production
  instruct models. At 20\% flag rate (right), seven of nine cells land
  in the \downstreammin--\downstreammax\% band, close to the language-modeling ceiling
  (\cref{tab:flagging_cross_scale}). Phi-3
  Mini on SQuAD and MedQA are the two exceptions, task-model
  interactions rather than method floors. All downstream runs use each
  model's WikiText-trained probe applied without task-specific
  training. Task-specific random-ranker baselines are not computed
  (see \cref{sec:appendix_downstream}); rates reflect non-overlapping
  catches relative to confidence, not excess over a downstream-specific
  null.}
\label{fig:downstream_ceiling}
\end{figure}

\FloatBarrier
\medskip
Across recipes, architecture and training recipe bound the upstream
ceiling on activation monitorability: frozen monitors can only
exploit the signals training preserved.

\section{Related work}
\label{sec:related_work}

Prior work on activation probes, hidden-state monitors, output-side
uncertainty, and monitorability usually begins with a trained model
and asks what signal can be extracted from its hidden states,
generated text, or output probabilities. This paper asks an upstream
question: whether architecture and training preserved the signal a
monitor would need. Decision quality is the first target; the
broader object is target-specific observability, where the question
is which safety- or reliability-relevant signals remain internally
readable after training. This framing connects probing validity,
behavioral activation directions, output-side uncertainty,
deployment-time monitoring, and model-design work on architecture
and training dynamics.

\medskip
\textbf{Probing validity and output-independent signals.}
Linear probes were introduced by \citet{alain2017understanding} to
monitor intermediate representations. Probing has a known gap
between what a probe finds and what the model
uses~\citep{belinkov2022probing}. \citet{hewitt2019designing}
address one side with control tasks: selectivity measures whether a
probe finds more than it could from random linguistic labels on the
same representation. A probe can pass this test and still be
redundant with the model's own output confidence. Partial
correlation with confidence covariates reframes the probing-validity
question from ``more than random?''\ to ``more than the output
already provides?'' On GPT-2~124M, raw Spearman is $\rawspearman$,
selective but half confidence-redundant ($\stdcontrol$ after
covariates). Across \nabsorbmodels{} models with control-sensitivity
data, controlling for confidence removes
$\confabsorbmean\% \pm \confabsorbstd\%$ of the raw probe signal,
ranging $\confabsorbmin$--$\confabsorbmax\%$ across the cohort
(\cref{sec:signal}). Hidden states encode answer correctness before
generation completes~\citep{zhang2025reasoning}, and
chain-of-thought success is predictable from pre-completion
representations~\citep{afzal2025knowing}; both works evaluate
discrimination between internal representations and correctness
labels without controlling for output confidence, so the
output-independent component of their reported signals is
unmeasured. Output-side readouts sharpen the same distinction:
$\ocresid$ asks whether the surviving signal remains after a
trained final-layer predictor has been given access to the output
representation, so the comparison is not merely probe versus
confidence but mid-layer monitor signal versus tested output-side
readout. Our scope also extends to variation that fixed-model
studies do not surface: the confidence-controlled signal survives
on six families and collapses on Llama~3B and 8B under the same
protocol, and checkpoint dynamics show the collapse is
training-emergent rather than a static property of the converged
model.

\medskip
\textbf{Truthfulness, factuality, hallucination, and behavioral directions.}
Linear directions for truthfulness have been found via unsupervised
consistency~\citep{burns2023discovering}, geometric
structure~\citep{marks2024geometry}, cross-model
universality~\citep{burger2024truth}, and inference-time causal
intervention~\citep{li2023inference}. \citet{zou2023representation}
read linear directions for behavioral concepts from intermediate
layers. \citet{han2025simple} train lightweight factuality probes
on hidden states for long-form generation, achieving competitive
hallucination detection with a single forward pass.
\citet{azaria2023internal} train classifiers on activations to
detect model-generated falsehoods.
\citet{goldowskydill2025detecting} use linear probes to detect
strategic deception from activations, finding high AUC
(0.96--0.999) but noting the probe sometimes fires for the topic of
deception rather than genuine intent. These methods target semantic
or behavioral labels and evaluate by discrimination metrics. Our
observer targets a different object (confidence-residual per-token
loss) and measures a different property (architecture-dependent
preservation vs.\ collapse); none of these works test
cross-architecture variation or training-emergent collapse.

\medskip
\textbf{Output-side uncertainty, calibration, and selective prediction.}
Output-level methods operate on what the model has already exposed.
Calibrated-confidence ranking~\citep{guo2017calibration} captures
most of the operationally useful signal; the
$\catchfloor$--$\catchceiling\%$ exclusive error rate at 10\% flag
rate is the residual it leaves uncovered. Scalar monotone
recalibrations preserve Spearman ranks; a nonlinear MLP on
$[\conf, \anorm]$ absorbs no additional signal
($\pcorr = \nonlincontrol$ vs.\ $\stdcontrol$), so nonlinear
miscalibration is not the residual's source. Language models can
estimate their own accuracy in some
settings~\citep{kadavath2022language}, but the output-side predictor
on the full last-layer representation still leaves $\ocresid > 0$
at every tested GPT-2 scale. Verbal uncertainty is governed by a
single linear feature in representation
space~\citep{ji2025calibrating}, correlating only moderately with
semantic uncertainty; verbal assertiveness correlates with output
confidence, so this feature may overlap with what our covariates
remove. \citet{kuhn2023semantic} cluster sampled generations by
meaning across multiple forward passes;
\citet{kossen2025semantic} approximate semantic entropy from hidden
states via linear probes; both predict an output-level quantity.
Output-level retrieval-augmented generation (RAG) detectors also
operate on generated text: NLI
faithfulness~\citep{honovich2022true}, sample
consistency~\citep{manakul2023selfcheckgpt}, and atomic-claim
factual precision~\citep{min2023factscore} all target what the
model already exposed. Selective prediction methods including early
exit~\citep{schuster2022confident} and verbalized uncertainty for
LLMs~\citep{wen2025know} make exit decisions on output-level
signals. The observer catches a non-overlapping error class:
tokens where the model is confident and the output gives no warning
but the mid-layer activations encode the failure. On SQuAD~2.0,
MedQA-USMLE, and TruthfulQA, a WikiText-trained probe catches
$\downstreammin$--$\downstreammax\%$ of errors exclusively at 20\%
flag rate in seven of nine model-task cells. Output-level and
observer methods are complementary because they ask different
questions: what the output reveals versus what the activations
preserve.

\medskip
\textbf{Activation monitoring at scale, monitorability, and evasion.}
Production probes are fragile across
distributions~\citep{kramar2026building}. Kram\'{a}r et al.\ build
production-ready activation probes for a model whose representation
encodes the relevant signal; this paper identifies the upstream
condition: whether the architecture-recipe choice preserved that signal in
the first place. Rule-based activation
monitoring~\citep{rozenfeld2026gavel}, dynamic safety
monitoring~\citep{oldfield2026beyond}, and monitorability
metrics~\citep{guan2025monitoring} measure monitoring quality
across deployed models without isolating which upstream choices
determine signal preservation. Mechanistic interpretability
findings require causal methods and separation of correlation from
causation~\citep{sharkey2025open}; \citet{mcguinness2025neural}
show that models can learn to evade activation monitors, and
\citet{jian2025metacognitive} find that LLMs have emergent capacity
to monitor and modulate their own activation patterns. The
observer direction is near-orthogonal to dominant variance on
GPT-2~124M (PC1 cosine $\pcaOneCosine$), suggesting that
decision-quality signals can occupy low-variance subspaces that
variance-dominant decompositions and some evasion analyses may
miss.

\medskip
\textbf{Architecture, scaling, and training as signal-preservation axes.}
Model-design work usually treats architecture and training recipe
as controls for loss, compute efficiency, scale, latency, and
downstream capability. Controlled training suites such as
Pythia~\citep{biderman2023pythia} make it possible to ask how these
choices affect training dynamics rather than only final task
scores. This work adds a different dependent variable to that
design space: whether training preserves monitorable internal
signals. In Pythia, a decision-quality signal appears early in both
matched-width configurations, but one configuration preserves it
while another erases it as perplexity improves. Across production
families, recipes shift which configurations collapse. Observability
therefore connects activation monitoring to architecture search and
signal engineering (\cref{def:signal_engineering}): the question is
not only how to read a frozen representation, but how to design
training so that oversight-relevant signals are preserved or
amplified as readable from frozen activations.

Chain-of-thought monitorability preservation~\citep{korbak2025cot,
xiong2026free} addresses a parallel axis on a different signal:
the legibility of reasoning traces under training pressure. The two
axes are independent: a model could have monitorable
chain-of-thought and erased activation signals, or vice versa. This
paper's contribution is the activation-level question with
controlled-recipe evidence.

\medskip
Across these literatures, the common assumption is that the
trained model is fixed and the problem is to extract, calibrate,
harden, or compare whatever signal remains. Our results move the
question upstream. For decision quality, and potentially for other
safety and reliability targets, the relevant signal may be formed
during training and later preserved, compressed, scattered, or
erased. Observability therefore turns activation monitoring into a
model-design problem: architecture and training determine what
oversight can see, and signal preservation becomes a training-time
engineering target.

\section{Discussion}
\label{sec:discussion}

\textbf{Training shapes monitorability.}
Training does not merely determine what a model predicts. It
determines which internal evidence remains available to oversight.
In Pythia's controlled training, both matched-width configurations
form the decision-quality signal early, but training preserves it
in one configuration and erases it in another while perplexity
improves. Predictive optimization and observability outcomes are separable
consequences of architecture and training. Capability and monitorability
are not inherently in tension: the healthy 16-layer, 8-head Pythia
configuration preserves both through convergence, while the 24-layer,
16-head class loses observability. Across
independent recipes the collapse point changes: Llama collapses at
a different configuration (1B healthy, 3B and 8B collapsed), and
Mistral~7B v0.3 and Llama~3.1 8B share the same architecture
(32-layer, 32-head, 4096-hidden) yet differ in observability
(\cref{sec:architecture}). Across these recipes, loss does not
predict observability.

\medskip
\textbf{Monitoring ceilings are set at training time.}
Activation monitoring treats the trained model as fixed, but the
model may already lack the signal. In collapsed
configurations, full layer sweeps, swept nonlinear probes,
token-budget checks, and tested output-side readouts all fail to
lift observability above the detection floor. The dynamics show why: training
actively erases a signal the architecture initially formed. Once
architecture and training are fixed, monitors are bounded by the
observability of the trained model. Scale and probe-width sweeps
confirm the bound (\cref{tab:cross_family_scaling},
\cref{sec:signal}). Frozen monitors read signals that training
preserves or amplifies.

\medskip
\textbf{Observability as a design axis.}
Observability has two operational uses. The immediate use is
model selection: a lab evaluating trained candidates can measure
observability before deployment, alongside loss and capability. The deeper use is design. Architecture and
training-recipe choices jointly bound whether monitorable signals
are preserved, amplified, compressed, or erased through training. This is an
upstream property of the trained model, and it bounds what any
frozen monitor can read.

The downstream catch results show why the design axis matters
(\cref{fig:downstream_ceiling}). At 20\% flag rate, seven of nine
model-task cells across SQuAD, MedQA, and TruthfulQA fall in the
$\downstreammin$--$\downstreammax$\% exclusive-catch band, with the
observer catching about one in seven of \qwen{}~7B Instruct's MedQA
errors that confidence does not flag. These results show the
operational stakes of preserving or amplifying signal at training
time.

\medskip

\textbf{From monitoring to signal engineering.}
Once a model is frozen at a collapsed configuration, no tested
probe or output-side readout recovers the signal.
Checkpoint dynamics show the signal was not impossible to represent
in that architecture: it appeared early and was later erased. The
architecture had the capacity, and training did not preserve it. This
opens an intervention window between formation and erasure, and
makes monitorability a design problem, not only a diagnostic one.
Signal engineering (\cref{def:signal_engineering}) names this
design problem. Decision quality is the first demonstrated target.
The paper establishes the axis, not a signal engineering training
intervention.

Signal engineering has a geometric implication. On GPT-2~124M,
the observer direction is nearly orthogonal to dominant
representation variance (PC1 cosine $\pcaOneCosine$), and
reconstruction-trained SAEs partially obscure it. Whether the signal lands in readable geometry is itself a training
outcome.

\medskip
\textbf{Observability beyond decision quality.}
Decision quality is one instance of a protocol that generalizes.
For any target with specifiable labels and output-side readouts,
the same evidence hierarchy asks whether training preserves a
readable signal beyond output confidence. Changing the target
changes the labels and readouts, not the burden of evidence. The
same protocol applies (\cref{sec:method}). Candidate targets
include factuality, refusal appropriateness, retrieval faithfulness,
reward-model disagreement, tool-use reliability, and controlled
deception labels. Whether these targets exhibit the same
architecture-recipe dependence or collapse at the same
configurations remains open.

\medskip
\textbf{Limitations and future work.}
The paper's claims are tiered by evidence strength. Each tier
names its limitation and the advancing experiment.

\textit{Within-recipe causation.} Checkpoint dynamics provide the
primary within-recipe temporal evidence: both matched-width configurations form the
decision-quality signal early, then training erases it in the 24-layer,
16-head class while the 16-layer, 8-head configuration recovers, at
comparable final perplexity. Within the Pythia suite, depth and
head count co-vary at the (24-layer, 16-head) class, so isolating which axis
drives the divergence requires single-axis controlled training. Why one configuration recovers from the mid-training dip
while the other does not is open. \emph{Advancing the claim:
single-axis controlled training at matched scale, and finer-grained
checkpoint analysis during the divergence window.}

\textit{Observational replication across recipes.} Cross-family
evidence and the Llama cliff show the pattern replicates under
different training recipes. Cross-recipe causation is the next
test, requiring controlled training across recipes. \emph{Advancing
the claim: controlled training at matched scale across recipes.}

\textit{Operational transfer.} Operational utility depends on
task-model factors beyond token-level observability. The next
bound is direct overlap with single-pass hallucination
detectors~\citep{kuhn2023semantic,kossen2025semantic} on the
confidence-invisible subset. \emph{Advancing the claim: direct
comparison with semantic-entropy variants on the
confidence-invisible subset.}

\textit{Method scope.} A separate residualizer-fit split preserves
the observability regime on \residSplitNmodels{} representative
models (\cref{sec:appendix_residualizer_split}). The tested
output-side capacity bottleneck is not the explanation for
$\ocresid$. Stronger or qualitatively different output-level
monitors remain outside scope. Adaptive evasion against the trained
observer~\citep{mcguinness2025neural} is the next threat-model
study. This paper measures the upstream signal availability that
any monitor depends on. For decision-quality or per-token error
targets, results reported without confidence covariates may
substantially overstate the output-independent component: in our
model suite, controlling for confidence removes $\confabsorbmean$\% of raw
signal on average across \nabsorbmodels{} models. \emph{Advancing
the claim: tests against stronger output-side readouts and adaptive
evasion studies.}

\textit{Scale.} The largest model tested is \qwen{}~32B. Whether
observability persists at frontier scale is unknown, although
\qwen{}~0.5B--32B and GPT-2~124M--1.5B show no degradation across
the tested within-family ranges. \emph{Advancing the claim:
observability measurements at frontier scale, particularly under
training-recipe variations adopted at frontier compute budgets.}

\medskip
Architecture and training shape the geometry of the residual
stream, determining whether the decision-quality signal remains
readable or scatters beyond the reach of activation monitors.
Activation monitorability is decided upstream, before monitor
design begins. Training sets the observability ceiling.

\section{Conclusion}
\label{sec:conclusion}

Architecture decides whether activation monitoring catches the
confident errors output-confidence monitoring misses. Training
decides what remains readable, before any monitor is built. In Pythia's controlled
training, both matched-width configurations form the
decision-quality signal early. Training then preserves it in Pythia
1B (16-layer, 8-head) and erases it in Pythia 1.4B
(24-layer, 16-head). Both improve in perplexity, and the 1.4B reaches lower
final perplexity than the 1B.
Predictive capability and monitorability are not inherently in tension.
The output-controlled residual collapses with observability. Across
independent recipes, the collapse point shifts but the same split
between preservation and collapse recurs.

Monitorability is an upstream design axis, alongside loss and
capability. It is measurable on the trained model and shaped through
architecture and training-recipe choices that preserve readable
decision-quality signals. Decision
quality is the first demonstrated target. The approach generalizes
to other safety and reliability targets. Monitoring is bounded by
the observability of the trained model. The next problem is signal
engineering: training models not only to predict well, but to
preserve or amplify monitorable signals in activations. Architecture
sets the conditions for observability, and training determines what
remains readable.

\section*{Reproducibility and disclosure}
\label{sec:reproducibility}

\paragraph{Reproducibility.}
Code version v5.1.0 (commit
\texttt{e0b505ed6e7aab76e6b3d510972b16396097cb02}) is tagged at
{\small\url{https://github.com/tmcarmichael/nn-observability/tree/v5.1.0}}.
The codebase and results are archived at Zenodo under the concept DOI
\href{https://doi.org/10.5281/zenodo.19435674}{10.5281/zenodo.19435674},
which resolves to the latest version (the v5.1.0 snapshot at the time
of submission).
Per-dataset documentation lives in \texttt{DATA.md} at the code release.
Committed results JSONs are the source of truth; each paper-scope
result includes a provenance block recording the model revision,
generating script, timestamp, and compute device. Hugging Face model
revisions are pinned in \texttt{results/model\_revisions.json}, and
dataset revisions in \texttt{results/dataset\_revisions.json}. Every
entry in \texttt{model\_revisions.json} is programmatically verified
against the Hugging Face API; the latest report under
\texttt{results/manifest\_verification/} records the verification
timestamp and exact-SHA-match status per entry, and is regeneratable
via \texttt{python scripts/verify\_manifest\_revisions.py}. The full
result-file inventory is published as a Croissant~1.1 metadata
descriptor at \texttt{croissant.json}, validated against the official
MLCommons spec by \texttt{just check-croissant}.
From a clean clone, \texttt{uv sync --frozen --extra transformer}
prepares the environment. \textit{Verification (no GPU, seconds).}
\texttt{uv run pytest tests/ -q \&\& just check} validates every
committed result JSON against its schema, every manifest entry against
the Hugging Face API, every paper-cited macro against its source files
and key paths, and every scope against
\texttt{analysis/load\_results.py}. \textit{Full regeneration
(single H100/A100, hours-to-days).} Two recipes drive the bulk:
\texttt{just pythia-suite} and \texttt{just downstream-all}.
Cross-family and other experiments have their own script. Software:
Python~3.12, PyTorch~2.8, and Hugging Face Transformers, managed
with uv.

\paragraph{LLM disclosure.}
Claude (Anthropic) assisted with Python scripts for GPU data
collection and statistical analysis, experiment notebooks, prose
editing, and section-level consistency checks against the results
JSONs. ChatGPT (OpenAI) was used for review and proofreading.

\paragraph{Broader impact.}
Architecture and training recipe bound whether activation
monitoring has a signal to exploit on a deployed model.
Monitorability is partially decided at training time, not bolted on
later. In high-stakes domains such as medical QA and
context-grounded generation, observability is a pre-deployment
measurement alongside latency and accuracy. The observer flags a
specific class of confidence-invisible failures. It does not detect
fluent factual errors, the boundary demonstrated by TruthfulQA, and
has not been tested against adaptive attacks.
\citet{mcguinness2025neural} demonstrate that activation monitors
can be evaded, and the same threat model applies here: the signal
does not ride a dominant variance direction (PC1 cosine
$\pcaOneCosine$). Deployment should treat activation monitoring as
complementary to output-level checks, not as a replacement.

\bibliographystyle{plainnat}
\Urlmuskip=0mu plus 1mu\relax
{\sloppy\hypersetup{urlcolor=refblue}\bibliography{references}}

\clearpage
\appendix
\crefalias{section}{appendix}
\crefalias{subsection}{appendix}
\crefname{appendix}{Appendix}{Appendices}
\Crefname{appendix}{Appendix}{Appendices}
\section*{Appendix}

\section{Statistical hardening}
\label{sec:appendix_hardening}

This appendix and the three that follow supplement
\cref{sec:method,sec:signal}. Together they cover the methodology
hardening package: statistical hardening (this section), layer and
split validation (\cref{sec:appendix_validation}), confidence-control
audits (\cref{sec:appendix_confidence}), and signal characterization
(\cref{sec:appendix_geometry}).

\subsection{20-seed statistical hardening}

20 independent observer heads at layer~11 of GPT-2~124M (seeds 42--61):
$\pcorr = \hardpcorr \pm \hardstd$, 95\% CI $[\hardlo, \hardhi]$, per-seed
range $[\hardseedmin, \hardseedmax]$, seed agreement $\hardsagree$. Layer~11 is
the peak layer under the 20-seed hardening protocol; the main 7-seed cross-family
protocol peaks at layer~8 with $\corepcorr$, consistent with the $\pm \topthreeflatness$ top-3
flatness reported in \cref{sec:signal}. This CI covers probe initialization
variance on fixed activations and a fixed dataset. It does not capture
uncertainty from dataset, checkpoint, or model family variation.

\subsection{Qwen 14B seed distribution}

\qwen{}~14B shows a bimodal seed distribution
($\sagree = \qFourteensagree$): per-seed values cluster near
$\qFourteenclusterlo$ (four seeds) or $\qFourteenclusterhi$ (three
seeds), and the reported mean ($\qFourteenpcorr$) averages across both
clusters. The scaling claim (\cref{sec:architecture}) does not depend
on the exact value: every seed lands between $\qFourteenseedlo$ and
$\qFourteenseedhi$, well above the detection floor and in line
with the rest of the Qwen family. The bimodality is preserved under
resampling; characterizing the geometric relationship between the
two clusters would require per-seed direction data we did not
collect.

\subsection{Supplementary statistical tests}

\textbf{Variance decomposition.}
A three-level decomposition attributes $\varfamily\%$ of the variance
to family membership, with three singleton-cell families inflating the
between-family share by construction; the matched 3B pair
(\cref{sec:architecture_stats}) and the Pythia controlled suite
(\cref{sec:pythia}) are the inferential tests.

\textbf{Uncertainty on $\eta^2$.}
A bootstrap over model means (10{,}000 resamples with replacement, $n =
\nmodels$) gives a 95\% CI of $[\etasqlo, \etasqhi]$ on $\eta^2$, with
the upper bound at the ceiling. The point estimate is close to the
maximum because within-family variance is small relative to between-family
variance at matched scale; the wide CI reflects the $n = \nmodels$
sample and the singleton-cell structure above, not instability of the
measurement.

\textbf{Leave-one-family-out sensitivity.}
Removing any single family keeps the permutation test significant
($p \leq \loofmaxp$ in all cases; weakest exclusion: without
\loofweakestfamily{}, $p = \loofweakestp$; without Gemma, $p = \loofgemma$;
without Llama, $p = \loofllama$). The result is not driven by any
single family.

\textbf{Jonckheere-Terpstra within-\qwen{} trend.}
Extending the within-\qwen{} sweep to 32B yields a non-monotonic
six-point trajectory ($\qHalfpcorr$, $\qOneFivepcorr$, $\qThreepcorr$,
$\qSevenpcorr$, $\qFourteenpcorr$, $\qThirtyTwopcorr$). The
Jonckheere-Terpstra test does not support a monotonic direction
($p = \jtp$; $\jtpct\%$ of cross-group comparisons decline, near the
$50\%$ chance baseline), and $\pcorr$ is bounded between $\qwenrangelo$
and $\qwenrangehi$ across $64\times$ scale. The cross-family
separation in \cref{sec:architecture_stats} carries the inferential
weight; the within-family scaling story is null. The mixed-effects
scale coefficient ($p = \scalep$) confirms within-family variation
does not explain cross-family differences.

\textbf{Supplementary ANCOVA.}
ANCOVA with scale and family yields family
$F(\ancovaFamdf) = \ancovaFamF$, $p = \ancovaFamp$;
$\log_{\text{params}}$ $F(\ancovaScaledf) = \ancovaScaleF$,
$p = \ancovaScalep$. Per-seed observations are not independent (shared data
and layer selection), so these $p$-values are anticonservative.
The mixed-effects model is the primary test because its random
intercepts per model absorb within-model correlation from shared data
and layer selection.

\textbf{Within-Pythia statistical tests.}
The 8-configuration permutation test (\cref{sec:pythia}) is exact:
$\binom{8}{2} = 28$ possible 2-vs-6 partitions, and the
(24-layer, 16-head) labeling is the unique partition producing the
largest healthy-minus-collapsed gap in $\pcorr$ ($\pythiaobsgap$), so
$p = \pythiapermp$. The healthy mean is $\pythiahealthymean$ ($n = 6$,
range $\pythiaTwoEightpcorr$ to $\pythiaOneSixtypcorr$); the collapsed
mean is $\pythiacollmean$ ($n = 2$, range $\pythiaFourTenpcorr$ to
$\pythiaOneFourpcorr$). Two supplementary checks: a Mann-Whitney U
between the two groups yields $U = 0$, exact one-sided
$p = \pythiapermp$; leave-one-configuration-out across the 8 sizes
preserves
$\max(\text{collapsed}) < \min(\text{healthy})$ on every removal.
The standing gap $\pythialoomingap$ is set by Pythia~2.8B (lowest
healthy) against 1.4B (highest collapsed); LOO removals only widen it.
Treating 1.4B-deduped as a third replication at the
collapse configuration, the 9-point 3-vs-6 exact permutation p-value
is $\pythiapermpdedup$ ($1/84$); the three replications range from
$\pythiatrilo$ to $\pythiatrihi$ (width $\pythiatriwidth$) despite
$3.5\times$ variation in parameters and two Pile
corpora.

\textbf{Selective prediction.}
Downstream exclusive catch rates for the observer (WikiText-trained,
applied without task-specific training) on SQuAD~2.0, MedQA-USMLE,
and TruthfulQA are reported in \cref{tab:downstream_tasks}.
Confidence alone has higher single-signal precision at every tested
flag rate; the observer's value is the non-overlapping catch
(\cref{sec:architecture}).

\subsection{Equivalence test (TOST) procedure}
\label{sec:appendix_tost}

The matched-hyperparameter MLP-vs-linear comparison
(\cref{sec:signal}) uses two one-sided
tests~\citep{lakens2017equivalence} under Schuirmann's
intersection-union framing. Per model $i$, the seed-averaged
paired delta $\bar{\Delta}_i$ is the difference in $\pcorr$
between the MLP probe and the linear probe. Across
$n = \tostNmodels$ models, let
$\bar{\Delta}$ denote the pooled mean and
$\mathrm{SE}(\bar{\Delta}) = s_\Delta / \sqrt{n}$ the standard
error, with $s_\Delta$ the unbiased sample standard deviation
($\mathrm{ddof} = 1$). Two one-sample $t$-statistics with $n - 1$
degrees of freedom test
\begin{align*}
H_0^{\mathrm{lower}}\!&:\,\mu_\Delta \leq -\tostMargin
  \quad \text{vs.}\quad \mu_\Delta > -\tostMargin, \\
H_0^{\mathrm{upper}}\!&:\,\mu_\Delta \geq +\tostMargin
  \quad \text{vs.}\quad \mu_\Delta < +\tostMargin.
\end{align*}
Equivalence at family-wise $\alpha = 0.05$ is declared when
$\max(p_{\mathrm{lower}}, p_{\mathrm{upper}}) < 0.05$. This maximum
is the reported TOST $p$-value. Equivalently, the $90\%$ CI on
$\mu_\Delta$ lies inside $\pm \tostMargin$. The margin is set to
twice the empirical top-three layer-flatness tolerance
($\tostMargin = 2 \cdot \topthreeflatness$).

\section{Layer and split validation}
\label{sec:appendix_validation}

\subsection{Layer selection and test-split confirmation}

Layer selection is two-stage. A seed-42 sweep produces a candidate
set: the top four layers by seed-42 $\pcorr$, depth-filtered to
$\le 80\%$ of total depth, plus the seed-42 argmax. A 7-seed
evaluation (seeds 43--49) at each candidate selects the reported
layer by argmax of the 7-seed mean; reported $\pcorr$ is the
7-seed distribution at that layer. Selection over the small
candidate set introduces bounded optimism: layer profiles are
smooth across the top 40--60\% of depth, and across tested models
the top three candidate layers differ by less than
$\topthreeflatness$ in $\pcorr$: Mistral~7B top-3 range
$\topthreeMistral$, Phi-3~Mini $\topthreePhi$, \qwen{}~14B
$\topthreeqFourteen$, \qwen{}~3B $\topthreeqThree$, Llama~1B
$\topthreellamaOne$, GPT-2~124M $\topthreegptS$.
Selecting any of the top three layers in place of the reported peak
would change no model's $\pcorr$ by more than $\topthreeflatness$ and
no cross-family comparison by more than $\xfamCompTol$.

A separate three-seed test split evaluates the same layer on
non-overlapping probe initializations. Across 12 cross-family
models with test splits, the mean absolute test-vs-validation gap
is $\testvalmeanabs\%$ (max $\testvalmaxabs\%$; \qwen{}~1.5B). Nine of 12
models show test $<$ validation; the remaining 3 show test $>$
validation by at most $\testvalposmax$ absolute.
The largest gap (\qwen{}~1.5B: $\qOneFivetest$ test vs.\
$\qOneFivepcorr$ validation, $\qOneFivetestgap\%$) is consistent with
the smaller sample ($n = 3$ vs.\ $n = 7$): subsampling the validation
seeds into groups of three reproduces comparable variance. Cross-family
rankings are identical on test data (\qwen{}~3B $\qThreetest$ vs.\
Llama~3B $\llamapcorr$, a $\llamatestgap\times$ gap), and
within-family flatness is preserved (\qwen{} test means span
$\qwentestlo$ to $\qwentesthi$ across 0.5B--32B; \qwen{}~14B has no
test split).

\subsection{Separate residualizer-fit split}
\label{sec:appendix_residualizer_split}

The standard protocol fits the OLS residualizer on the same documents
used to train the probe. To test whether this in-sample fit creates
an artifact in the residual target, we construct a disjoint
residualizer-fit pool R and apply R-fit coefficients without refit to
the probe-training pool T and the evaluation pool V. The three pools
draw disjoint WikiText-103 documents at the same hidden-dim token
budget as the canonical protocol; the residualizer is a three-coefficient
OLS fit (softmax, activation norm, intercept), and the probe is the
same linear head trained on the residualized binary target on T.
Per-model diagnostics including per-seed values, $\beta$ coefficients,
token counts, and pos-label fractions appear in the
\texttt{*\_residualizer-split.json} files in \texttt{results/}.

We run this on \residSplitNmodels{} models: GPT-2~124M (healthy
anchor), Pythia~1B (healthy controlled), Pythia~1.4B (controlled
collapse), Llama~3.2 3B (observational collapse). The pre-specified
pass criterion is regime preservation: healthy models satisfy
$\pcorr \geq \pythiahealthyminR$, collapsed models
$\pcorr \leq \detectionfloor$, with $\pm 0.02$ tolerance
(\cref{tab:residualizer_split}).

\begin{table}[h]
\centering
\small
\caption{Residualizer-fit split robustness. Reference $\pcorr$ is the
canonical 7-seed mean. Split-fit $\pcorr$ uses an OLS residualizer
fit on a disjoint document pool. All four models satisfy the
pre-specified criteria. Max direct-baseline
$|\Delta\pcorr| = \residSplitMaxDelta$ across the four models; the
canonical-reference difference for Pythia~1.4B is $0.033$ because the
direct collection path gives $0.127$ rather than the all-layer-sweep
reference $\residPyOneFourRef$ (both below the detection
floor).\protect\footnotemark}
\label{tab:residualizer_split}
\begin{tabular}{llrrcc}
\toprule
Model & Regime & Reference $\pcorr$ & Split-fit $\pcorr$ & Criterion & Result \\
\midrule
GPT-2~124M    & healthy anchor         & $\residGptRef$        & $\residGptSplit$        & $\geq \pythiahealthyminR$ & pass \\
Pythia~1B     & healthy controlled     & $\residPyOneRef$      & $\residPyOneSplit$      & $\geq \pythiahealthyminR$ & pass \\
Pythia~1.4B   & controlled collapse    & $\residPyOneFourRef$  & $\residPyOneFourSplit$  & $\leq \detectionfloor$ & pass \\
Llama~3.2 3B  & observational collapse & $\residLlamaRef$      & $\residLlamaSplit$      & $\leq \detectionfloor$ & pass \\
\bottomrule
\end{tabular}
\end{table}
\footnotetext{For Pythia~1.4B, the same-pool baseline recomputed
under the direct-collection path is $0.127$ versus the canonical
all-layer-sweep reference value $\residPyOneFourRef$. Both lie below
the detection floor. The $|\Delta\pcorr|$ statistic uses
the direct baseline so that the residualizer split is isolated from
the collection path.}

The matched within-Pythia pair is the strongest cell. Pythia~1B
(16-layer, 8-head) and Pythia~1.4B (24-layer, 16-head) share recipe,
training data, hidden dimension, and probe-token budget; the only
difference is the configuration tuple. Under split-fit, the
distinction holds: 1B remains healthy at $\residPyOneSplit$, 1.4B
remains collapsed at $\residPyOneFourSplit$. This matched pair makes
an in-sample residualizer-fit artifact unlikely: the same split-fit
protocol preserves the 1B/1.4B separation under disjoint
residualizer-fit pools.

The OLS coefficients fit on T versus R are nearly identical across
all four models. The softmax coefficient varies by at most
$\olsCoefDelta$ from its base magnitude (the four base values cluster
near $-5$); the activation-norm coefficient and the intercept are
essentially unchanged. Positive-label fractions on T differ between
$\beta_T$ and $\beta_R$ by less than $\posLabelTol$ absolute in every
model. The
residualizer is a deterministic function of token statistics, not a
knob whose fitting pool meaningfully changes the target distribution.

\section{Confidence-control audits}
\label{sec:appendix_confidence}

\subsection{Control sensitivity}

On GPT-2~124M, raw Spearman of $\rawspearman$ drops to $\softmaxonly$
after softmax-only control, holds at $\stdcontrol$ after adding norm,
and reaches $\nonlincontrol$ under a nonlinear MLP control on
$[\conf, \anorm]$. Softmax is the main confound; norm adds nothing
beyond softmax; the residual survives nonlinear deconfounding.
Adding logit entropy as a third covariate absorbs
$\entropyabsorbpct\%$ on top of the standard covariates ($\entropycontrol$),
indicating the observer partially reads the shape of the output
distribution, not just the peak.

\begin{table}[ht!]
\centering
\caption{Control sensitivity across three architectures. Softmax
dominates absorption in all three; standard controls remove between
39\% (Pythia~410M) and 63\% (\qwen{}~7B) of the raw signal; norm-only
does not remove the residual; nonlinear controls leave similar or
slightly higher residuals than standard controls; entropy removes
additional output-shape information.}
\label{tab:control_sensitivity_xfamily}
\begin{tabular}{lccc}
\toprule
Control & GPT-2~124M & \qwen{}~7B base & Pythia~410M (collapse) \\
\midrule
None (raw Spearman)       & $\rawspearman$       & $\qSevenCtrlNone$    & $\pyFourTenCtrlNone$    \\
Norm only                 & $\normonly$          & $\qSevenCtrlNorm$    & $\pyFourTenCtrlNorm$    \\
Softmax only              & $\softmaxonly$       & $\qSevenCtrlSoftmax$ & $\pyFourTenCtrlSoftmax$ \\
Standard (softmax + norm) & $\stdcontrol$        & $\qSevenCtrlStd$     & $\pyFourTenCtrlStd$     \\
Nonlinear MLP             & $\nonlincontrol$     & $\qSevenCtrlNonlin$  & $\pyFourTenCtrlNonlin$  \\
Standard + logit entropy  & $\entropycontrol$    & $\qSevenCtrlEntropy$ & $\pyFourTenCtrlEntropy$ \\
\bottomrule
\end{tabular}
\end{table}

The control ordering is consistent across architectures. \cref{tab:control_sensitivity_xfamily}
extends the GPT-2~124M analysis to a healthy cross-family model
(\qwen{}~7B base) and to a within-Pythia collapse point (Pythia~410M).
In all three, softmax dominates absorption and standard (softmax +
norm) covariates remove between 39\% and 63\% of the raw signal.
Norm-only absorbs almost nothing on top, and the nonlinear MLP
control yields a similar residual to the linear standard control. The collapse point shows the same control
ordering at a shifted absolute scale.

\subsection{Verifying ``\confabsorbmean\% is confidence''}

The fraction of raw signal removed by controlling for confidence
covariates varies across families but is consistently large. Across \nabsorbmodels{}
models in \nfamilies{} families, standard covariates (max-softmax $+$
activation norm) absorb $\confabsorbmean\% \pm \confabsorbstd\%$ of
the raw Spearman correlation, ranging
$\confabsorbmin$--$\confabsorbmax\%$ across the cohort. The pattern
is consistent: confidence is the dominant confound on every
architecture tested. The ``half to three-quarters'' summary in the
main text reflects this range.

\textbf{Absorption gradient tracks observability.} Absorbed fraction
tracks architecture-dependent observability. Healthy models span
$\confabsorbmin\%$--$64.5\%$, with GPT-2~124M at the low end and
\qwen{}~32B at the top. Both Llama collapse points concentrate at the
upper end: Llama~3.2 3B at $63.8\%$ and Llama~3.1 8B at
$\confabsorbmax\%$, the highest in the cohort: what little raw
correlation exists there is almost entirely confidence-redundant. The direction of the
gradient, higher absorption where the confidence-independent signal
is smaller, is the operational reading of the unified claim:
architecture-dependent observability and output-independence are the
same property.

\subsection{Target circularity}

We construct the binary target by regressing per-token loss on
$[\conf, \anorm]$ via OLS and thresholding the residual at zero.
\Cref{sec:signal} presents the complete six-test validity argument;
we expand on four of those tests here:

\begin{enumerate}
  \item Hand-designed readouts of the same activations fail to predict the
    same target (\cref{sec:signal}),
    so the target is not trivially recoverable from activation geometry.

  \item The probe succeeds on some architectures (\qwen{}~3B $\qThreepcorr$) and fails
    on others (Llama~3B $\llamapcorr$) under the identical target definition.
    The variation is in the representations, not the target formula.

  \item On \qwen{}~7B, within-domain C4 training fails ($\qSevenCfourWithin$; Llama~8B
    also fails at $\llamaEightCfourWithin$), so the target does not manufacture signal on all
    activation geometries even within architectures that succeed on WikiText.

  \item Seed agreement on the residualized (confidence-removed)
    observer scores remains $\residsagree$, so the cross-seed
    consensus is on the confidence-independent component rather than
    on a shared confidence pathway.
\end{enumerate}

\section{Signal characterization}
\label{sec:appendix_geometry}

\subsection{Signal composition (Shapley decomposition)}

Four named controls account for about $\shaptotalpct\%$ of the raw
signal under a Shapley value decomposition, computed across all 24
orderings of the 4 controls. The 4-control Shapley total is not
directly comparable to the 2-control (max-softmax $+$ activation
norm) absorption range reported in \cref{sec:signal} because the
control sets differ. The controls are: (i) max-softmax
confidence, (ii) logit entropy, (iii) geometric typicality (Mahalanobis
distance of the activation from the training-set activation mean under
the training covariance), and (iv) token frequency (log unigram
frequency computed on the probe training corpus). The remaining $\shapresidpct\%$ resists all tested controls. Individual
component attributions vary by ordering (confidence ranges from
$\shapconfmin\%$ to $\shapconfmax\%$, entropy from $\shapentmin\%$ to
$\shapentmax\%$) because the covariates are correlated, but the total
explained fraction and the unexplained residual are stable. Input-side controls
(bigram surprisal) do not absorb the signal.

\subsection{Signal geometry}

The observer direction is nearly orthogonal to the dominant variance
(PC1 cosine $\pcaOneCosine$, top 10 PCs capture $\pcaTopTenPct\%$).
The signal reads from the low-variance subspace of the residual
stream. Standard dimensionality reduction would miss it. As one
example, a $\saefeatures$-feature reconstruction-trained sparse
autoencoder~\citep{bricken2023monosemanticity} probe ($\saepcorr$)
underperforms the raw linear observer ($\corepcorr$) despite
$32\times$ more input dimensions, with rank correlation
$\saerankcorr$ between the two. A sparse decomposition optimized for
reconstruction concentrates on high-variance directions; the observer
direction lies in a low-variance subspace that variance-dominant and
reconstruction-trained decompositions partially obscure.

\subsection{GPT-2 within-family scaling}

\begin{table}[t]
\centering
\caption{GPT-2 scaling under matched 350 ex/dim, 7-seed protocol. Partial
correlation is stable across 12$\times$ scale (\gptpcorrLo--\gptpcorrHi).
The output-controlled residual $\ocresid$ is positive at every size
(\gptOcMin--\gptOcMax), so the confidence-residual signal is not fully
recovered by the tested final-layer predictor at any scale. The
output-controlled fraction $\ocresid/\pcorr$ varies between
\gptOcFracMin\% and \gptOcFracMax\% without a monotonic trend in size;
predictor-capacity sensitivity is addressed cross-family by the Qwen
7B output-side width sweep (\cref{sec:signal}).}
\label{tab:gpt2_scaling}
\begin{tabular}{lccccccc}
\toprule
Model & Params & Peak layer & $\pcorr$ & $\pm$ std & $\ocresid$ & $\ocresid / \pcorr$ & $\sagree$ \\
\midrule
GPT-2          & 124M  & L8 (67\%)   & $0.288$ & $0.001$ & $0.140$ & 49\% & $0.952$ \\
GPT-2~Medium   & 355M  & L17 (71\%)  & $0.290$ & $0.004$ & $0.127$ & 44\% & $0.940$ \\
GPT-2~Large    & 774M  & L25 (69\%)  & $0.291$ & $0.004$ & $0.151$ & 52\% & $0.943$ \\
GPT-2~XL       & 1558M & L36 (75\%)  & $0.296$ & $0.017$ & $0.084$ & 28\% & $0.829$ \\
\bottomrule
\end{tabular}
\end{table}

Across the four GPT-2 sizes (124M, 355M, 774M, 1.5B), partial
correlation lands in $\gptpcorrLo$--$\gptpcorrHi$ and the
output-controlled residual in $\gptOcMin$--$\gptOcMax$, with
output-controlled fraction $\ocresid/\pcorr$ in
$\gptOcFracMin\%$--$\gptOcFracMax\%$ (\cref{tab:gpt2_scaling}).
All four are evaluated under the canonical 350 ex/dim, 7-seed
protocol used throughout the cross-family suite. The within-family
scaling test pairs with the \qwen{}~0.5B--32B sweep
(\cref{sec:architecture}) under a different architecture template.

\section{Checkpoint dynamics}
\label{sec:appendix_dynamics}

\cref{tab:checkpoint_dynamics} reports the full checkpoint trajectory
for the two matched-width Pythia configurations ($d = 2048$) analyzed
in \cref{sec:pythia}. Both configurations produce healthy $\pcorr$
and positive $\ocresid$ at step~256 ($\approx$0.5B tokens). The 1.4B
(24L/16H) configuration peaks at step~1{,}000 ($\pcorr = \pythiaOneFourMidpcorr$,
$\ocresid = \pythiaOneFourMidOC$), then declines. The 1B (16L/8H) configuration
follows a similar early trajectory, dips below the detection floor at
step~16{,}000, and recovers by step~32{,}000. Both configurations
converge to comparable perplexity (18--21) at the final checkpoint.
Seed agreement dips and recovers in both configurations; the 1.4B
ends at $\sagree = 0.947$ alongside collapsed $\pcorr = \pythiaOneFourpcorr$. Peak layer trajectories differ between configurations. The 1B starts
at L12 (steps~256--1k), shifts to L8--L10 across most of mid-training
(returning briefly to L12 at the step-16k dip), and converges to L10.
The 1.4B starts at L17--L18, drifts to L10--L13 mid-training, and
returns to L17 as the signal collapses.

\begin{table}[h]
\centering
\caption{Checkpoint dynamics for two matched-width Pythia configurations
($d = 2048$). Both form healthy observability at the earliest measured
checkpoint. The (16L/8H) configuration recovers after a mid-training
dip; the (24L/16H) configuration does not, converging near the
detection floor despite comparable perplexity.}
\label{tab:checkpoint_dynamics}
\small
\setlength{\tabcolsep}{3.5pt}
\begin{tabular}{llrlrS[table-format=1.3]S[table-format=1.3]S[table-format=1.3]lS[table-format=1.3]}
\toprule
Model & Config & Step & Tokens & Ppl & {$\pcorr$} & {$\pm$ std} & {$\ocresid$} & Peak & {$\sagree$} \\
\midrule
Pythia~1B    & 16L/8H & 256 & 537M & $1424.9$ & 0.315 & 0.001 & 0.079 & L12 (75\%) & 0.979 \\
Pythia~1B    & 16L/8H & 1k & 2.1B & $139.4$ & 0.303 & 0.003 & 0.099 & L12 (75\%) & 0.970 \\
Pythia~1B    & 16L/8H & 2k & 4.2B & $62.6$ & 0.288 & 0.001 & 0.119 & L10 (62\%) & 0.968 \\
Pythia~1B    & 16L/8H & 4k & 8.4B & $41.0$ & 0.226 & 0.006 & 0.097 & L8 (50\%) & 0.951 \\
Pythia~1B    & 16L/8H & 8k & 16.8B & $32.3$ & 0.242 & 0.013 & 0.094 & L10 (62\%) & 0.939 \\
Pythia~1B    & 16L/8H & 16k & 33.6B & $27.8$ & 0.134 & 0.008 & 0.023 & L12 (75\%) & 0.887 \\
Pythia~1B    & 16L/8H & 32k & 67.1B & $25.1$ & 0.252 & 0.017 & 0.122 & L10 (62\%) & 0.932 \\
Pythia~1B    & 16L/8H & 64k & 134.2B & $22.9$ & 0.257 & 0.015 & 0.133 & L10 (62\%) & 0.951 \\
Pythia~1B    & 16L/8H & 128k & 268.4B & $20.4$ & 0.206 & 0.012 & 0.085 & L10 (62\%) & 0.967 \\
Pythia~1B    & 16L/8H & 143k & 299.9B & $20.6$ & 0.246 & 0.012 & 0.120 & L10 (62\%) & 0.974 \\
\midrule
Pythia~1.4B  & 24L/16H & 256 & 537M & $1660.9$ & 0.297 & 0.002 & 0.065 & L17 (71\%) & 0.973 \\
Pythia~1.4B  & 24L/16H & 1k & 2.1B & $136.3$ & 0.314 & 0.001 & 0.113 & L18 (75\%) & 0.977 \\
Pythia~1.4B  & 24L/16H & 2k & 4.2B & $61.8$ & 0.231 & 0.002 & 0.057 & L13 (54\%) & 0.970 \\
Pythia~1.4B  & 24L/16H & 4k & 8.4B & $40.3$ & 0.254 & 0.004 & 0.098 & L10 (42\%) & 0.963 \\
Pythia~1.4B  & 24L/16H & 8k & 16.8B & $30.4$ & 0.156 & 0.007 & 0.037 & L11 (46\%) & 0.940 \\
Pythia~1.4B  & 24L/16H & 16k & 33.6B & $25.4$ & 0.123 & 0.008 & 0.019 & L13 (54\%) & 0.914 \\
Pythia~1.4B  & 24L/16H & 32k & 67.1B & $22.5$ & 0.113 & 0.003 & 0.011 & L17 (71\%) & 0.893 \\
Pythia~1.4B  & 24L/16H & 64k & 134.2B & $20.6$ & 0.214 & 0.007 & 0.092 & L13 (54\%) & 0.942 \\
Pythia~1.4B  & 24L/16H & 128k & 268.4B & $18.4$ & 0.125 & 0.004 & 0.013 & L17 (71\%) & 0.949 \\
Pythia~1.4B  & 24L/16H & 143k & 299.9B & $18.4$ & 0.106 & 0.006 & 0.003 & L17 (71\%) & 0.947 \\
\bottomrule
\end{tabular}
\end{table}

\FloatBarrier
\section{Token budget sensitivity}
\label{sec:appendix_token_budget}

Cross-scale comparisons require matching token budgets to hidden
dimension. Without this, larger models appear to have weaker signal
because their higher-dimensional representations need proportionally
more training examples.

\begin{figure}[t]
\centering
\includegraphics[width=0.7\textwidth,alt={Seven token budgets from 150
  to 1000 ex/dim showing partial correlation at peak layer for Qwen
  0.5B. Signal is flat near 0.15 below 450 ex/dim, rises sharply to
  0.215 at 600, and continues to 0.248 at 1000.}]{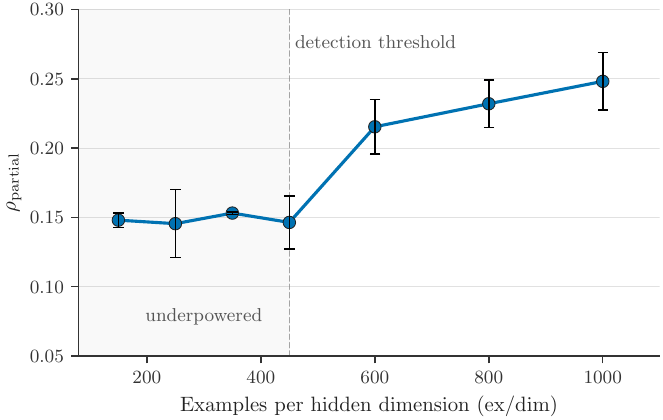}
\caption{Ex/dim sensitivity at \qwen{}~0.5B. Seven token budgets
  showing $\pcorr$ (7-seed mean) at peak layer. The signal is below
  the detection threshold between 150 and 450~ex/dim and rises
  sharply between 450 and 600.}
\label{fig:exdim}
\end{figure}

At \qwen{}~0.5B ($d = 896$), a seven-point sweep from 150 to 1,000
examples per hidden dimension (ex/dim) reveals a detection threshold
between 450 and 600 ex/dim
(\cref{fig:exdim}). Below 450, $\pcorr$ is flat at $\exdimbelowlo$ to $\exdimbelowhi$
(7-seed means). At 150 ex/dim the seed std is $\exdimSeedStdLow$, so the
measurement is precise; the signal is weak because the probe is
under-resourced, not because it is noisy. At 600, $\pcorr$ rises to $\qHalfpcorr$, the value
reported in \cref{tab:cross_family_scaling}, and continues to
$\exdimeight$ at 800 and $\exdimthousand$ at 1{,}000.

The threshold implies that small models ($d < 1{,}000$) require
more data per dimension than larger models do to produce stable
probes. All cross-scale comparisons in this paper use ex/dim ratios
above the detection threshold for each model's hidden dimension. An
earlier measurement of \qwen{}~0.5B at 190~ex/dim ($\exdimOneNinety$)
was below the detection threshold. The cross-family table reports
the 600~ex/dim result ($\qHalfpcorr$), which clears the detection
threshold; the endpoint remains mildly budget-sensitive between 600
and 1{,}000~ex/dim ($\qHalfpcorr$ to $\exdimthousand$).

On \qwen{}~14B, increasing from 68 to 350~ex/dim on the same model
raises $\pcorr$ from $\budgetvone$ to $\budgetvthree$.

The ex/dim metric scales the training set with hidden dimension, so
absolute token budgets grow with model size. At the reported ratios,
the smallest cross-family model (GPT-2~124M, $d = 768$) uses
$\sim$270{,}000 training tokens and the largest hidden dimension
($d = 5120$, used by \qwen{}~14B and 32B) uses $\sim$1.8M. Collapsed
models are not bottlenecked by absolute budget: Llama~3B
($\llamapcorr$) and 8B ($\llamaEightpcorr$) both run with absolute
budgets above $10^6$ tokens and still collapse. The within-family
comparisons (\qwen{}~0.5B--32B, Llama~1B--8B, Pythia~70M--12B) hold
ex/dim above each model's detection threshold.
The threshold's dependence on absolute budget is out of scope.

\section{Cross-domain transfer}
\label{sec:appendix_cross_domain}

\subsection{The probe is portable but the calibration data is not}

A WikiText-trained observer transfers to C4 web text across five
families (Qwen, Llama, Gemma, Mistral, Phi), retaining
$\cfourretainmin$--$\cfourretainmax\%$ of the WikiText signal
(\cref{tab:cross_domain}). The probe reads the same kind of quality signal on
text it was never trained on.

\begin{table}[h]
\centering
\caption{Cross-domain transfer: WikiText-trained probe evaluated on
  C4 (transfer) vs.\ probe trained directly on C4 (within-domain).
  Percentages show C4 transfer as a fraction of WikiText signal.}
\label{tab:cross_domain}
\small
\begin{tabular}{lccc}
\toprule
Model & WikiText $\pcorr$ & C4 transfer & C4 within-domain \\
\midrule
\qwen{}~3B       & $0.263$ & $0.190$ (72\%) & $-0.021$ \\
\qwen{}~7B       & $0.255$ & $0.191$ (75\%) & $-0.029$ \\
Llama~1B         & $0.286$ & $0.159$ (55\%) & $-0.046$ \\
Llama~3B         & $0.091$ & $0.041$ (45\%) & $-0.033$ \\
Mistral~7B       & $0.313$ & $0.155$ (50\%) & $-0.010$ \\
Phi-3~Mini       & $0.300$ & $0.192$ (64\%) & $0.223$ \\
Gemma~3~1B       & $0.216$ & $0.151$ (70\%) & $0.047$ \\
\bottomrule
\multicolumn{4}{l}{\footnotesize Percentages computed from unrounded values.} \\
\end{tabular}
\end{table}

The asymmetry is consistent: WikiText$\to$C4 transfer works, but
C4$\to$C4 within-domain training fails on most tested models, with
Phi-3~Mini the exception ($\phiCfourWithin$). The probe is portable
across distributions, but training requires calibrated data. C4 is
noisier and more heterogeneous than WikiText, so the OLS
residualization against confidence produces a noisier binary target.
On most families the probe does not converge on a stable direction
under these conditions, even with the same total token count. We do
not have a predictor for which families succeed on within-domain C4
training.

\subsection{Code transfer}

On GPT-2~124M, a WikiText-trained probe transfers to
CodeSearchNet~\citep{husain2019codesearchnet}
Python at $\pcorr = \codetransfer$, exceeding the WikiText source
signal ($\corepcorr$). Transfer to OpenWebText~\citep{gokaslan2019openwebtext} is weak
($\owttransfer$), weaker than the C4 transfer pattern. The code
result suggests the decision-quality signal encodes structural prediction
difficulty (closing brackets, completing function signatures, matching
indentation) that generalizes across natural language and code. Code
has cleaner syntactic structure than web text, which may explain why
the probe finds more signal there than on its training distribution.

\subsection{Low-observability models}

Llama~3B shows weak signal on both WikiText ($\llamaThreeWiki$) and
C4 transfer ($\llamaThreeCfour$). Llama~1B shows the same
asymmetry as high-observability models: WikiText $\llamaOneWiki$, C4
transfer $\llamaOneCfour$, C4 within-domain $\llamaOneCfourWithin$.
Low observability at 3B is a property of
the model's representations, not the input domain.

\section{Downstream evaluation}
\label{sec:appendix_downstream}

\subsection{Tasks and protocol}

The same 20\% ceiling recurs in seven of nine cells across three
downstream tasks: SQuAD~2.0~\citep{rajpurkar2018know},
MedQA-USMLE~\citep{jin2021disease}, and
TruthfulQA~\citep{lin2022truthfulqa}, measured on three production
instruct models (\qwen{}~7B Instruct, Mistral~7B Instruct v0.3, Phi-3
Mini Instruct), each using its own WikiText-trained probe applied
without task-specific training (\cref{tab:downstream_tasks}).
SQuAD~2.0 uses the first $n = 1{,}000$ answerable questions in
dataset order from the validation split with provided context
(no retrieval); MedQA-USMLE uses the first $n = 1{,}000$ items in
dataset order from the four-option test split
scored by exact option match; TruthfulQA uses the multiple-choice
validation split (817 questions, MC1 single-correct format) scored
by normalized substring match against the MC1 correct option. Each model generates an answer via
greedy decoding with a minimal task prompt. The observer score and
max-softmax confidence are averaged over generated tokens (excluding
the prompt) to produce per-question scores. Questions are ranked by
each score and flagged at the specified rate. A question is an
exclusive observer catch if the observer flags it, confidence does
not, and the answer is wrong.

\begin{table}[ht]
\centering
\caption{Exclusive catch rates across three production instruct models
  and three downstream tasks, using each model's WikiText-trained probe
  applied without task-specific training. Columns report catch rate at
  10\% and 20\% flag rate (percentage of errors caught exclusively by
  the observer). Seven of nine cells at 20\% fall between $10.9\%$ and
  $13.4\%$; the two exceptions (Phi-3~Mini Instruct on SQuAD at $6.6\%$
  and on MedQA at $4.1\%$) are task-model interactions discussed
  below. Confidence has higher single-signal precision
  at every flag rate; these are the observer's non-overlapping catches.}
\label{tab:downstream_tasks}
\small
\begin{tabular}{lcccccccc}
\toprule
 & \multicolumn{2}{c}{\qwen{}~7B-I} & & \multicolumn{2}{c}{Mistral~7B-I} & & \multicolumn{2}{c}{Phi-3~Mini-I} \\
\cmidrule(lr){2-3} \cmidrule(lr){5-6} \cmidrule(lr){8-9}
Task & @10\% & @20\% & & @10\% & @20\% & & @10\% & @20\% \\
\midrule
SQuAD~2.0          & 5.0\% & 10.9\% & & 9.7\% & 11.0\% & & 5.8\% & 6.6\% \\
MedQA-USMLE        & 9.4\% & 13.4\% & & 6.5\% & 11.2\% & & 1.0\% & 4.1\% \\
TruthfulQA         & 7.9\% & 12.7\% & & 6.7\% & 10.9\% & & 7.7\% & 13.1\% \\
\bottomrule
\end{tabular}
\end{table}

\subsection{Per-cell breakdown}

Seven of the nine model-task cells at 20\% flag rate fall between
$\downstreammin\%$ and $\downstreammax\%$, near the language-modeling
ceiling band ($\catchsaturation\%$ at 20\%,
\cref{tab:flagging_cross_scale}). Two cells sit below: Phi-3~Mini
Instruct on SQuAD ($6.6\%$) and MedQA ($4.1\%$). Phi-3~Mini Instruct
preserves observability on WikiText ($\phipcorr$) and on TruthfulQA
($13.1\%$ at 20\%), so the SQuAD and MedQA outcomes are task-model
interactions rather than a floor on the method. The observer direction
learned from Wikipedia next-token prediction reads the same signal
across reading comprehension, medical, and factual QA
(\cref{fig:downstream_ceiling}). On MedQA with \qwen{}~7B Instruct,
among the \medqaerrorpct\% of questions the model gets wrong, at
20\% flag rate about one in seven is confidently wrong and caught
only by the observer.

\textbf{Random-baseline note.} A task-specific random-ranker baseline
analogous to the GPT-2 language-modeling baseline
(\cref{sec:architecture}) is not computed for the downstream cells
in this paper. The language-modeling analog is $\randbaselineten\%$
at 10\% flag rate and $\randbaselinetwenty\%$ at 20\%, derived from
the disagreement of two rankers with fixed flag rates. The
downstream catch rates above ($\downstreammin$--$\downstreammax\%$
at 20\%) sit near this analog, so the operational claim is the
non-overlapping catch of a frozen WikiText probe, not a margin over
random. Computing a downstream-specific random-ranker baseline is
left for future work.

\subsection{TruthfulQA boundary}

The catches above operate over all errors at a fixed flag rate.
Among the narrower subset of confidently wrong answers, where the
model asserts a fluent falsehood with high confidence, the observer
does not discriminate
(area under the ROC curve, AUC $\tqaAUCqwen$ on \qwen{}~7B-I, $\tqaAUCmistral$ on Mistral
7B-I, $\tqaAUCphi$ on Phi-3~Mini-I, all near chance). The observer
catches token-level prediction failures at the aggregate rate; it
does not catch fluent reproduction of learned falsehoods, and the
three-model consistency defines that boundary as a property of the
method, not of the model.

\section{Instruction-tuning robustness}
\label{sec:appendix_instruct}

The cross-family pattern holds at every \qwen{} scale tested
(\cref{tab:instruct}):

\begin{table}[t]
\centering
\caption{Base vs.\ instruct across \qwen{} scales plus Mistral~7B.
  Instruction tuning preserves observability; changes are small relative
  to the cross-family spread.}
\label{tab:instruct}
\small
\begin{tabular}{lS[table-format=1.3]S[table-format=1.3]S[table-format=-1.3]}
\toprule
Scale & {Base $\pcorr$} & {Instruct $\pcorr$} & {$\Delta$} \\
\midrule
\qwen{}~0.5B   & \qHalfpcorr     & \qiHalfpcorr     & \qiHalfdelta \\
\qwen{}~1.5B   & \qOneFivepcorr  & \qiOneFivepcorr  & \qiOneFivedelta \\
\qwen{}~3B     & \qThreepcorr    & \qiThreepcorr    & \qiThreedelta \\
\qwen{}~7B     & \qSevenpcorr    & \qiSevenpcorr    & \qiSevendelta \\
\qwen{}~14B    & \qFourteenpcorr & \qiFourteenpcorr & \qiFourteendelta \\
Mistral~7B     & \mistralSevenpcorr & \mistralSevenIpcorr & \mistralSevenIdelta \\
\bottomrule
\multicolumn{4}{l}{\footnotesize Deltas computed from unrounded values.} \\
\end{tabular}
\end{table}

At 0.5B, instruction tuning increases the signal by
$\Delta = \qiHalfdelta$. At 1.5B through 14B, the delta is small
($\qiOneFivedelta$ to $\qiSevendelta$). The pattern extends beyond
\qwen{}: Llama~3.2 1B Instruct produces $\llamaOneipcorr$ versus
base $\llamaOnepcorr$ ($\Delta = \llamaOneidelta$); Mistral~7B
Instruct gives $\mistralSevenIpcorr$ against base
$\mistralSevenpcorr$ ($\Delta = \mistralSevenIdelta$). Whatever produces observability runs deeper than what instruction
tuning modifies.

\end{document}